\documentclass[]{interact}

\usepackage{epstopdf}% To incorporate .eps illustrations using PDFLaTeX, etc.
\usepackage[noabbrev, capitalise]{cleveref}
\usepackage{natbib}% Citation support using natbib.sty
\bibpunct[, ]{(}{)}{;}{a}{}{,}% Citation support using natbib.sty
\renewcommand\bibfont{\fontsize{10}{12}\selectfont}% Bibliography support using natbib.sty

\theoremstyle{plain}% Theorem-like structures provided by amsthm.sty

\theoremstyle{definition}

\theoremstyle{remark}

\usepackage{pifont}
\newcommand{\yes}{\textcolor{blue}{\ding{52}}} 
\newcommand{\no}{\textcolor{red}{\ding{56}}}

\usepackage[labelsep=space]{caption}
\usepackage{subcaption}
\usepackage{booktabs}

\usepackage{array}
\usepackage[usenames, dvipsnames]{xcolor}
\usepackage{multirow}

\begin{document}

% \articletype{ARTICLE TEMPLATE}

\title{Limited or Biased: Modeling Sub-Rational Human Investors in Financial Markets}

\author{
% \name{Anonymous Authors}
\name{Penghang Liu\textsuperscript{a}\thanks{CONTACT Penghang Liu. Email: penghang.liu@jpmchase.com}, Kshama Dwarakanath\textsuperscript{b}, Svitlana S Vyetrenko\textsuperscript{b}, and Tucker Balch\textsuperscript{a}}
\affil{\textsuperscript{a}J.P.Morgan AI Research, New York, New York, USA; \textsuperscript{b}J.P.Morgan AI Research, Palo Alto, California, USA}
}

\maketitle

\begin{abstract}
Human decision-making in real-life deviates significantly from the optimal decisions made by fully rational agents, primarily due to computational limitations or psychological biases. 
While existing studies in behavioral finance have discovered various aspects of human sub-rationality, there lacks a comprehensive framework to transfer these findings into an adaptive human model applicable across diverse financial market scenarios. In this study, we introduce a flexible model that incorporates five different aspects of human sub-rationality using reinforcement learning. 
Our model is trained using a high-fidelity multi-agent market simulator, which overcomes limitations associated with the scarcity of labeled data of individual investors.
We evaluate the behavior of sub-rational human investors using hand-crafted market scenarios and SHAP value analysis, showing that our model accurately reproduces the observations in the previous studies and reveals insights of the driving factors of human behavior. 
Finally, we explore the impact of sub-rationality on the investor's Profit and Loss (PnL) and market quality. Our experiments reveal that bounded-rational and prospect-biased human behaviors improve liquidity but diminish price efficiency, whereas human behavior influenced by myopia, optimism, and pessimism reduces market liquidity.
\end{abstract}


\section{Introduction}

In traditional economics and game theoretic studies, humans are often conceptualized as \textit{homo economicus} -- fully rational, self-interested agents with unbiased beliefs. However, empirical evidence has consistently demonstrated that real human behavior is far more intricate, and may not always align with perfect decision making (\cite{thaler1997effect, benartzi1999risk, chrisman2012variations}). 
Over the recent years, there has been a growing interest in behavioral economics in response to these complexities (\cite{thaler2016behavioral}). It attempts to incorporate insights from other social sciences, especially psychology, in order to enrich the standard economic models that fail to explain human behavior in real life.

We refer to such behavior as being \textit{sub-rational}, as opposed to perfectly rational decision-making. Many studies have been developed to reveal the various drivers of human sub-rationality, which can we categorized into the following two classes. First, human sub-rationality is \textbf{limited}. \cite{simon1955behavioral} suggested that humans may attempt to take sub-optimal decisions that are satisfactory rather than optimal due to limited access to information and processing power. The second facet of human sub-rationality is that humans are psychologically or cognitively \textbf{biased}. For example, \cite{kahneman1979prospect} developed prospect theory to explain human decision making under uncertainty. \cite{benartzi1995myopic,thaler1997effect} discussed that human decision-making can be myopic, with a strong emphasis on short-term gains/losses as opposed to those over longer terms. 

Despite the richness of discoveries in aspects of human sub-rationality, there lacks an effective and universal approach to model and predict human trading behavior influenced by these aspects in financial markets. 
Consequently, existing behavioral finance studies are mostly constrained on real market observations, and it remains unclear how each aspect of human sub-rationality will impact their investment strategy, profit, and the market dynamics. 
To fill this gap, we employ reinforcement learning (RL) to model market participants. The RL agents are trained in the market environment to learn a trading strategy that optimizes the specified reward function. In one of the first financial market applications of RL, \cite{nevmyvakarl} used RL to train an optimal algorithmic execution agent. Similarly, in subsequent work, \cite{spooner2018market,dwarakanath2021profit} use RL to design market makers that provide liquidity in the market. By default, the RL agent will obtain a trading strategy that is optimal in maximizing its cumulative rewards, upon sufficient training. Subsequently, we modify the Bellman equation to model each aspect of human sub-rationality as a deviation from these optimal decisions. 

In financial markets, labelled data that identifies market participants is typically proprietary, and it is often impossible to retrieve individual level trading activities of human investors (\cite{gutierrez2019mapping}). To address this limitation, we use a high-fidelity market simulator to design, train and test our RL agents. Research in finance is well facilitated by versatile market simulations, which provide feasible experiment control and concrete market observations (\cite{friedman2018double}).
Multi-agent market simulators have been applied in financial research to reproduce the scaling laws for returns, assess the benefits of co-location, investigate the impact of large orders, and evaluate trading strategies (\cite{lux1999scaling, byrd2019abides, balch2019evaluate}). 
These simulators promote the use of RL algorithms to learn complex trading strategies in a shielded, simulated environment before trying them out in real markets.
In particular, we employ ABIDES-Gym (\cite{byrd2019abides,amrouni2021abides}), which provides discrete event time based Discrete Event Multi-Agent Simulation (DEMAS) for financial markets. It allows us to simulate fine-grained intraday market dynamics, and assess the impacts of human trading behavior on market micro-structure based on the limit order book (LOB). In addition, we build probabilistic neural networks to model the biased internal beliefs of human investors that may differ from true market dynamics given by the simulator.

In this paper, we model and examine the behavior of sub-rational human investors in financial markets resulting from computational limitations or psychological biases.
We introduce five types of sub-rational human investors: bounded rational, psychologically myopic, prospect biased, optimistic, and pessimistic.
For each type of human investor, we demonstrate the corresponding trading strategy in a hand-crafted market scenario to intuitively explain the strategy. We also investigate the relationship between the investor's profits and loss (PnL) and their corresponding sub-rationality. 
In addition, we use an explainability tool called SHAP to investigate the major driving factors of human decision-making.
Last but not least, our experimental analysis discovers the impact of sub-rational investors on the market with regard to liquidity, volatility, and market efficiency. 

To the best of our knowledge, this is the first work that utilizes the RL framework to model human trading behavior in financial markets as caused by different aspects of sub-rationality. We summarize the main contributions of our work as follows:
\begin{itemize}
    \item We develop a comprehensive and adaptive RL framework to model human sub-rational trading behavior characterized by bounded rationality, myopia, prospect bias, optimism, and pessimism.
    \item To address the limited availability of real, individual trader data, we utilize market simulations to train and test the sub-rational RL agents. In addition, we build probabilistic neural networks to model the internal bias in human beliefs.
    \item We craft specific market scenarios and use SHAP analysis to deeply examine the different types of sub-rational trading behavior. Furthermore, we test the agents in controlled simulated markets to evaluate the impact of sub-rational behavior on market observables.
\end{itemize}

We believe our models will provide an effective framework to capture and examine human investor behavior while aiding in better understanding of their influence in financial markets.

\begin{table}[t]
\caption{Summary of sub-rational human trading behavior.} 
\centering 
\makebox[\textwidth]{
\small
\begin{tabular}{|l|c|c|c|c|c|}
\hline
& Bounded  & Myopic & Prospect Biased & Pessimistic & Optimistic \\ \hline 
\multirow{ 2}{*}{Method} & Boltzmann & temporal & biased reward & biased internal &  biased internal \\ 
& softmax & discounting & \& internal model & model & model \\ \hline
\multirow{ 2}{*}{Behavior} & random error & aligned with short-& risk-averse in gains & small inventory & large inventory \\ 
& in decisions & term momentum & risk-seeking in losses & infrequent trading & aggressive buy/sell\\ \hline
Deciding & \multirow{ 2}{*}{N/A} & inventory & \multirow{ 2}{*}{inventory} & \multirow{ 2}{*}{N/A} & \multirow{ 2}{*}{N/A} \\ 
Factors & & short-term momentum & & & \\ \hline
PnL Variance & high & low & high & low & high \\ \hline
Improve Liquidity & \yes & \no & \yes & \no & \no \\ \hline
Reduce Volatility & \yes & \no & \no & \no & \no \\ \hline
Increase Efficiency & \no & \yes & \no & \yes & \no \\ \hline
\end{tabular}
}
\label{tab:summary}
\end{table}

\section{Literature Review}
\subsection{Human Sub-rationality}
One of the most important topics in behavioral finance is to understand and model the decision-making process of humans in real life. 
\cite{simon1955behavioral} was one of the first critics of modeling economic agents as having unlimited information and processing capabilities. He introduced the idea of ``bounded rationality'' to describe a more realistic conception of human problem solving capabilities. 
Lots of the departures from rational choice can be captured by the prospect theory proposed by \cite{kahneman1979prospect}, which provides a purely descriptive theory of human decision-making under uncertainty.
\cite{ainslie1975specious,thaler1981economic} discussed the importance of self-control in human behavior, where temporal discounting is often utilized to explain human's myopic preference for short-term results rather than long-term results.
Researchers have also discovered other aspects of human sub-rationality as resulting from biased internal beliefs such as optimism (\cite{sharot2007neural}) and illusion of control (\cite{barber2002online, song2013overconfidence}). 
While we believe that humans are far too complicated to be represented by any mathematical model, the focus of our paper is to develop a unified framework to model the decision-making of humans influenced by different types of sub-rationality discovered in previous studies.

Due to the limitation in data availability and realistic human models, there lacks analytical studies of the impact of human sub-rationality on financial markets. Meanwhile, regulatory literature has examined the impact of electronic traders as compared to that of human traders in markets.
\cite{algo_trading} investigated the impact of algorithmic trading on equity market quality measured in terms of quoted spread, price efficiency and volatility. 
They observed that more algorithmic trading lead to narrower spreads, better price efficiency but higher volatility based on real data, with the effects differing based on asset size.
\cite{woodward2017need} investigated the market impact of electronic traders that feature high frequency trading (HFT) techniques, and provided insights to control it from the perspective of regulators.
Note that these studies examine the influence of algorithmic and high frequency trading techniques, which are used by a specialized class of electronic investors. In this work, we compare the impact of sub-rational human investors with that of perfectly rational electronic investors (with no change in trading frequency) in financial markets.

\subsection{Human Decision Models}
Reinforcement learning (RL) is an effective approach to obtain a decision-maker's policy for different rewards/incentives in specific environments. 
Theoretical efforts have been made to model the human decision-making process using RL. A related field is inverse reinforcement learning (IRL), which aims to bypass the need for reward design by inferring the reward from observed human demonstrations. 
There are numerous hypotheses behind sub-rationality of humans. 
\cite{raja2001towards} attributed human sub-rationality to constraints on resources and address a meta-level control problem for a resource-bounded rational human.
% Raja et al. \cite{raja2001towards} address the meta-level control problem for a resource-bounded rational agent.
% Wen et al. \cite{wen2021modelling} introduced generalized recursive reasoning (GR2) as a novel framework to model \textcolor{blue}{humans.}%agents.
\cite{evans2015learning, evans2016learning} modeled structured deviations from optimality with different hierarchical levels of rationality when inferring preferences and beliefs.
% Evans et al. \cite{evans2015learning, evans2016learning} modeled structured deviations from optimality when inferring preferences and beliefs with different hierarchical levels of rationality. 
In a recent work, \cite{Human_irrationality21} investigated the effects of human irrationality on reward inference. They introduced a framework to describe different aspects of human irrationality using the Bellman equation by modifying the max operator, the transition function, the reward function, and the sum between reward and future value.
While literature in IRL encompasses various interesting models of humans, the goal is to infer the reward function from real human demonstrations.
However, in financial markets, it is rarely possible to acquire historical trading data of individual, human investors.
% in the financial market it is usually not applicable to directly acquire the historical data of real human trading activities.
% , and provided theoretical as well as empirical analysis of the effects of different types of irrationality on reward inference.}
% While literature in IRL encompasses various interesting models of humans, the goal is to infer human beliefs from demonstrations rather than to design an agent that strategizes like a human. 
The goal of our work is to use behavioral models for human traders to model an investor agent that trades like a human in financial markets, and to subsequently analyze human traders' impact on the market in a simulated environment.
% The goal of our work instead is to use observations about human trading activity and to model an investor agent that trades like a human in financial markets, and then subsequently analyze human traders' impact on the market in a simulated environment.
% They are not primarily interested in investigating models of human decision-making, but instead focused on using such models to accurately infer human beliefs and preferences.

\subsection{Multi-agent Market Simulations}
% \penghang{market simulation, reinforcement learning, modeling human behavior, market impact}
Multi-agent simulators have become increasingly prevalent for modeling financial markets, which provide an alternative yet effective approach to study the market in rarely observed scenarios or with limitations of data.
% Market simulator has been a versatile tool for studying research questions in finance. 
\cite{lux1999scaling} introduced a multi-agent model of financial markets to support the time scaling law from mutual interactions of participants. In recent contributions, \cite{byrd2019abides} developed a discrete event simulator to investigate the market impact of a large market order. Additionally, \cite{vyetrenko2020get} proposed realism metrics to evaluate the fidelity of the simulated markets.
While these multi-agent market simulators can be populated with rule based trading agents, they allow for the use of reinforcement learning to develop agents that seek to optimize certain objectives.
% Reinforcement learning provides a new approach to utilize market simulation and define agents with different trading strategies. 
\cite{amrouni2021abides} wrapped market simulations in an OpenAI Gym framework facilitating the use of off-the-shelf RL algorithms, and train investors in various market environments.
% Amrouni et al. \cite{amrouni2021abides} employed the OpenAI Gym framework on market simulation, and train individual investor in different environments. 
\cite{dwarakanath2021profit} utilized RL to obtain the policy of market makers and subsequently investigated their impact on market equitability.
% Dwarakanath et al. \cite{dwarakanath2021profit} utilized RL to obtain the policy of market makers and investigated their impact on the market equitability.

\section{Background}
In this paper, we analyze and evaluate human investor behavior based on the market microstructure.
% In this work, we focus on the market microstructure analysis for evaluating and understanding human behavior. 
We now introduce the underlying concepts of market microstructure including the limit order book, and the corresponding market observables.

\begin{figure}[t]
    \centering
    \includegraphics[width=0.7\linewidth]{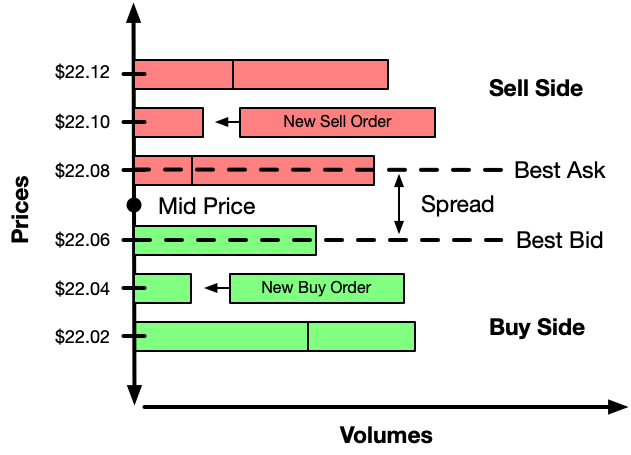}
    \caption{A snapshot of the LOB structure.}
    \label{fig:lob}
    \vspace{-2ex}
\end{figure}
\subsection{Limit Order Book (LOB) structure}
Due to the prevalence of high frequency trading, most of the financial market exchanges including NASDAQ and NYSE are now operating on an underlying dynamic mechanism known as the Limit Order Book (LOB) (\cite{rocsu2009dynamic, gould2013limit}). The LOB is a dynamic repository of ``buy"  and ``sell" limit orders, which provides a real-time granular view of the supply and demand at various price levels (\cite{bouchaud2002statistical,vyetrenko2020get}). 
\cref{fig:lob} provides a visual representation of the LOB structure. Here we summarize the key concepts of the LOB: 
\begin{itemize}
    \item \textbf{Limit Orders:} Unlike the traditional \textit{market orders} that represent the intentions to buy and sell immediately at the current market price, a \textit{limit order} specifies the minimum price that the trader intends to sell at, or the maximum price that the trader is willing to buy at, alongside the volume and direction (sell/buy).
    \item \textbf{Buy Side and Sell Side:} The LOB maintains limit orders on two sides, the buy and sell sides. The \textit{best bid} and \textit{best ask} represent the highest price a buyer is willing to pay and the lowest price a seller is willing to accept, respectively.
    \item  \textbf{Order Queue:} Orders from various market participants are collected and aggregated into queues based on their respective prices and time. The exchange matches buy and sell orders based on a first-in-first-out (FIFO) basis. 
\end{itemize}
The LOB offers a granular and accurate depiction of the underlying mechanisms governing modern financial markets. Moreover, the discrete nature of LOB offers a detailed examination of trader interactions, how liquidity is provided or consumed, and the evolution of order queues over time. Market simulations on LOB facilitate detailed behavioral analyses, enabling the exploration of how various trading strategies impact market dynamics.

\subsection{Metrics and Notations}\label{sec:metrics}
Here we describe the market metrics used for empirical analysis in our study. Following the notation in \cite{coletta2022learning}, we denote $p^i_a(t), v^i_a(t), p^i_b(t), v^i_b(t)$ as the quoted price and volume of ask and bid orders at $i$-th level of the LOB, which has $n$ levels on the ask side and $m$ levels on the bid side. We use the below metrics to describe the market dynamics in our study.
\begin{itemize}
    \item \textbf{Mid price} is the average of the best ask and best bid prices, i.e.,
    $$m(t) = \frac{p^1_a(t)+p^1_b(t)}{2}$$
    \item \textbf{Traded price} $p(t)$ is the price of the latest transaction before time $t$, which represents the most recent market price of the asset. 
    \item \textbf{Spread} is the difference between the best ask and best bid prices:
    $$\Delta(t) = p^1_a(t)-p^1_b(t)$$
    \item \textbf{Market depth} is the price difference between the worst ask and worst bid prices: 
    $$d(t) = p^n_a(t) - p^m_b(t)$$
    % \item \textbf{Order depth} is the difference between the order's limit price and the mid price.
    % \item \kshama{\textbf{Market depth} is the average of the price differences between the last, best ask orders and the best, last bid orders:
    % $$d(t)=\frac{\left(p_a^\infty(t)-p_a^1(t)\right)+\left(p_b^1(t)-p_b^\infty(t)\right)}{2}$$}
    \item \textbf{Traded Price Volatility} represents the degree of variation of the traded price series over the past $\delta$ time window:
    $$\sigma(t, \delta) = \text{standard deviation}(p(t-\delta), p(t-\delta+1), \dots, p(t))$$
    \item \textbf{Momentum} represents the trend of traded price over the over the past $\delta$ time window:
    $$\text{momentum}(t, \delta) = \frac{p(t)}{p(t-\delta)}$$
    \item \textbf{Volume imbalance} indicates the supply and demand inequality within the LOB:
    $$I(t) = \frac{\sum^m_{j=1}v^j_b(t)}{\sum^n_{i=1}v^i_a(t)+\sum^m_{j=1}v^j_b(t)}$$
    Volume imbalance is known to be an informative feature for trading as it can signal private information that can subsequently reduce market liquidity, and can impact the decisions of market makers thereby affecting market returns as well as execution agents \cite{chordia2002order,stoll2003market, zheng2012price}.
    
\end{itemize}
% \kshama{Minor: Spread and time window for volatility, momentum both use $\delta$ symbol.}
% \kshama{Should we describe the quoted and traded prices, volumes here? And why they are different.}

\subsection{Measures of Market Quality}\label{sec:quality_metrics}

Following the experiments of \cite{boehmer2021algorithmic}, we evaluate the market quality on three dimensions: liquidity, volatility, and information efficiency.

\subsubsection{Liquidity} 
Liquidity is a crucial aspect of financial markets, which indicates the ease of trading a security in the market without significantly impacting its market price. High liquidity markets typically have narrow bid-ask spreads and large traded volumes. We measure the liquidity of the stock based on the spread $\Delta(t)$, the relative effective spread RES, and the traded volume in every minute. 
The relative effective spread is calculated as:
\begin{align}
    \text{RES} = \frac{p(t)-m(t)}{m(t)}
\end{align}
where $p(t)$ and $m(t)$ are the traded price and mid-price at time $t$ respectively.
% $$
% \text{RES} = (\text{Price}_k - \text{MP}_k) / \text{MP}_k
% $$
% where Price$_k$ is the price of the $k$-th trade and MP$_k$ is the corresponding mid price at the time of the trade.
RES measures price changes associated with trading. 
The narrower the relative effective spread is, the more liquid the stock is.

\subsubsection{Volatility} 
Market volatility is a pivotal factor for market analysis, which can exert multifaceted influences on market quality through various channels. The prices of volatile assets have strong fluctuations and are often less predictable. Greater volatility tends to increase the cost of utilizing limit orders, making liquidity provision more expensive for market participants. As volatility rises, the likelihood of prices deviating significantly from the levels set in limit orders increases, making it more challenging for traders to execute orders at their desired prices. 
Define the price return over a time period $\delta$ as $\frac{p(t)}{p(t-\delta)}$.
We consider three measures of volatility: the standard deviation of 30-minute price returns (Std-30min-RET), the intraday price range (difference between the lowest and highest prices in a day), and the absolute value of daily returns ($|$RET$|$). 
Markets with low volatility have low values for the three volatility metrics. 

\subsubsection{Market efficiency}
According to the efficient market hypothesis (\cite{fama1970efficient,malkiel1989efficient}), market efficiency (also known as information efficiency or price efficiency) represents the degree to which available information is reflected in market prices. If market prices fail to fully encompass all available information, opportunities may arise for profit through the collection and processing of such information. An inefficient price also indicates that the markets are not doing an efficient job of allocating resources. We consider the autocorrelation of mid-price as an indicator of price efficiency following \cite{boehmer2009institutional}, which suggests that efficient pricing should follow a random walk and the autocorrelation should be close to 0. 
While it is infeasible to encompass all market information in the history in the real-world, in market simulations, we model and incorporate such information into the simulated fundamental value (see \ref{sec:simulation}). Consequently, our measurement of market efficiency extends to $|$TP-FP$|$ / FP, representing the relative disparity between the traded price and the fundamental price concurrently. A minimal separation signifies that the market operates with a high degree of information efficiency. 

\section{Reinforcement Learning for Investor Modeling}\label{sec:method}
In this section we describe how to model investors using reinforcement learning (RL). We choose to use the RL framework to model human investors due to the following reasons: (1) RL is more flexible compared to rule based approaches, and can adapt to different market scenarios. RL allows a more general representation of human behavior as arising from objective maximization rather than rigidly defined rule based behaviors. (2) RL offers a mathematical interpretation of different sub-rational human behaviors as deviations from the optimal policy, and provide controls of the degree of human sub-rationality. (3) RL models can be trained and tested in market simulations efficiently, which helps in evaluating the performance of different trading strategies. It also allows us to study diverse subrational behaviors without the need for real human demonstrations.

\subsection{Defining RL Investors}
In this work, we utilize reinforcement learning to obtain trading strategies for investors. Formally, we consider the Markov decision process (MDP) $(S, A, P_a, R_a)$ where $S$ and $A$ are the sets of states and actions. $P(s'|s,a)$ is the transition probability from state $s$ to $s'$ by taking the action $a$, and $R(s,a,s')$ is the immediate reward of the transition. The goal in RL is to maximize the expected sum of discounted rewards. 
% The value of the current state $s$ is therefore measured by the Bellman equation:
The Bellman equation relates the value of the current state $s$ to the one step reward and the value at the next state $s'$ as
\begin{align}
    V(s) = \underset{a}{\max} \underset{s' \in S}{\sum}{P(s'|s,a)\left(R(s,a,s') + \gamma V(s')\right)}\label{eq:bellman}
\end{align}
% \begin{equation}
%     V_{i+1}(s) = \underset{a}{\max} \underset{s' \in S}{\sum}{P(s'|s,a)(R(s,a,s') + \gamma V_i(s'))}
% \end{equation}
where $\gamma$ is the discount factor. 
We define a fully rational agent as one that picks the action that maximizes the right hand side of \cref{eq:bellman} when in state $s$. The corresponding optimal (rational) policy is
\begin{align}
    \pi(s) = \arg\underset{a}{\max} \underset{s' \in S}{\sum}{P(s'|s,a)\left(R(s,a,s') + \gamma V(s')\right)}\label{eq:rational_policy}
\end{align}

% {\color{green}SV: LOB needs to be described better for journal paper. perhaps use the chart that visualizes LOB (such as Figure 1 here https://arxiv.org/pdf/1906.02312.pdf). State and action space then need to be described more rigorously.
% }
% \kshama{Wondering if we want to say (rational electronic, subrational human) or just (rational, subrational) throughout?}
% \penghang{How about we say rational vs sub-rational in methods and models \cref{sec:method,sec:model}. And then in the market experiments \cref{sec:results} we say that we use rational RL agents for modeling electronic investors and sub-rational RL agents for modeling human investors.}

The RL agent represents an investor that tries to make profits by trading in the market. We model the rational investors and sub-rational human investors using the same state space, action space, and reward function.

\noindent\textbf{States:}
The states include the agent's observation of the market along with its internal states, including
%The states represent the RL investor agent's perception of the market and its current status, which includes
\begin{itemize}
    \item $\text{Cash}_t$: the amount of cash in the investor's account at time $t$. The agent starts the day with 1,000,000 cents.
    \item $\text{Holdings}_t$: the number of shares of the asset held by the agent at time $t$. The agent starts the day with no holding.
    \item $[\text{momentum}(t,1), \text{momentum}(t,10), \text{momentum}(t,30)]$: a vector reflecting the market momentum over the past 1, 10, and 30 minutes. 
    \item $\Delta(t)$: the market spread at time $t$.
    \item $d(t)$: the market depth at time $t$.
    \item $\sigma(t, 30)$: the traded price volatility in the past 30 minutes.
    \item Quote history: the information of quoted/placed orders in the past five minutes, including quoted prices and volumes.
    \item Trade history: the information of the executed trades in the past five minutes, including traded prices and volumes.
\end{itemize}
\noindent\textbf{Actions:}
The RL agent wakes up ever minute between 9:30am to 4:00pm, and takes an action defined by the following parameters
%The RL agents wake up ever minute between 9:30am to 4:00pm, and are allowed to take the following actions:
\begin{itemize}
    \item Direction: \{BUY, HOLD, SELL\}. In the case of BUY or SELL, the agent places a limit order of size 2.
    \item Limit order price (relative to the mid price): $\lbrace-0.5, -1, -1.5, -2\rbrace$ if the agent takes a BUY action, or $\lbrace+0.5, +1, +1.5, +2\rbrace$ if the agent takes a SELL action.
    \item Note that RL policy is formulated as a MDP, i.e., the transition from current state $s$ to the next state $s'$ only depends on the action $a$. We cancel all previous open orders of the RL agent before placing the new order from action $a$ so that the changes of inventory only come from the most recent orders.
    % \kshama{Same as order depth described in previous section?}
\end{itemize}
\noindent\textbf{Rewards:}
We define the one step reward as \begin{align}
    R(s_t,a,s_{t+1}) = \text{PortfolioValue}_t - \text{PortfolioValue}_{t-1}\label{eq:reward}
\end{align}
% $$R(s,a,s') = PortValue_t - PortValue_{t-1}$$
where $\text{PortfolioValue}_t = \text{Cash}_t + \text{Holdings}_t \times p(t)$ is the value of the investor's portfolio marked to the market, and $p(t)$ is the executed price of the last transaction before time $t$. 
% The intuition behind defining the reward function as in \cref{eq:reward} is as follows. 
Since the reward function measures the change of portfolio values in every minute, the cumulative reward over the horizon equals the agent's monetary profits at the end of the trading day (assuming there is no transaction cost and with temporal discounting $\gamma=1$). 
Here we formally define the profit and loss (PnL) at time $t$ as the change in portfolio value compared to the start of the day:
\begin{equation}
    \text{PnL} = \text{PortfolioValue}_{t} - \text{PortfolioValue}_{t_0}
\label{eq:pnl}
\end{equation}

\subsection{Training in Multi-Agent Market Simulations}\label{sec:simulation}
Market simulations are often employed to train RL agents as they offer a controlled environment to learn an optimal trading strategy in presence of reactive background trading agents, while also dealing with the inherent challenge of acquiring proprietary historical LOB data.
In this paper, we employ ABIDES-Gym, a discrete event multi-agent simulator that provides a high-fidelity market environment with thousands of trading agents, wrapped in an OpenAI Gym framework ~\cite{byrd2019abides,amrouni2021abides}. 
% Here, we describe the key features of ABIDES-Gym, and the definitions of RL agents and background agents in our simulated market.
% \subsubsection{Background Agents}
The simulated market contains the following background agents with different trading strategies and incentives. 
\begin{itemize}
    \item \textbf{Value Agents:} The value agents are designed to emulate fundamental traders who trade according to their belief of the exogenous value of an asset, i.e., \textit{fundamental price} \cite{wang12020046}. In this paper, we generate the time series of the fundamental price with a discrete-time mean-reverting Ornstein-Uhlenbeck process. Each value agent makes a noisy observation of the fundamental price, and places a sell order if the current mid price is higher than the observed fundamental price, or vice versa.
    \item \textbf{Market Maker Agent:} The market maker agent acts as a liquidity provider in the market, by placing limit orders on both sides of the LOB at regular intervals. The agent places equal volumes of liquidity at various price levels that are pre-defined with respect to the mid-price.
    %The agent repeatedly places orders on both side of the LOB with the order size split equally in to each level.
    \item \textbf{Momentum Agents:} The momentum agents trade based on the momentum of the asset price, by comparing the long-term average of the mid-price with the short-term average. Each agent places a buy order if the short-term average is higher, based on the belief that the price will increase in the future. On the other hand, if the long-term average is higher, the agent places a sell order.
    \item \textbf{Noise Agents:} The noise agents mimic retail traders that trade based on their own demand with no consideration of the LOB microstructure. 
    % They place orders in a random direction with a random size.
    Each noise agent arrives to the market at a random time, and places an order on a random side of the LOB.
\end{itemize}
We train the RL agent in the simulated market environment along with the background agents. As illustrated in \cref{fig:internal_model}, the market simulator receives the action of the RL agent at each step and provides the next state and reward information. The RL agent optimizes its trading policy based on the feedback from the environment.

\begin{table}[t]
    \centering
    \caption{The configuration of simulated markets. Except the exchange agent and market makers, the order sizes of the the background agents are randomly sampled from a heavy-tailed distribution with mean equals to 100. The actions of value agent follows a Poisson process with arrival rate $\lambda=1/5.7 \cdot 10^{-12}$ in nanoseconds, i.e., once per 3 minutes on average.}
    \begin{tabular}{|l|c|c|c|}
    \hline
         Agent Type & \# of Agents & Order Size & Action Frequency \\ \hline
         % Exchange Agent & 1 & N/A & N/A \\ \hline
         Value Agent & 2 & 100 & Poisson ($\lambda=1/5.7 \cdot 10^{-12}$) (ns) \\ \hline 
         Market Maker & 1 & 10 & Every minute\\ \hline
         Momentum Agent & 2 & 100 & Every 5 minutes\\ \hline
         Noise Agent & 20 & 100 & Once a day\\ \hline
    \end{tabular}
    \label{tab:configuration}
\end{table}

\subsection{Training on Internal Beliefs}\label{sec:internal_model}
Many human sub-optimal actions can be considered as the result of internal model error. That is to say that the reason that human actions might deviate from the optimal is that they have incorrect internal beliefs of the rules (dynamics) guiding how their actions affect the environment (\cite{reddy2018you}). Therefore, the agent obtains a biased policy which is sub-optimal in the real world but near-optimal with respect to its internal model of the dynamics. In reinforcement learning, the environment dynamics are captured by the transition probability $P(s'|s,a)$ in \cref{eq:bellman}, which represents the probability of the next state $s'$ given the current state $s$ and action $a$. A sub-rational human has a internal model which gives biased estimations of the transition probability. For example, the optimistic/pessimistic agent in \cref{sec:opti_pessi} has biased estimations of positive outcomes over the negative outcomes. The prospect biased agent in \cref{sec:prospect} overestimates the probability of unlikely events.

\begin{figure*}[t]
    \centering
    \includegraphics[width=1\linewidth]{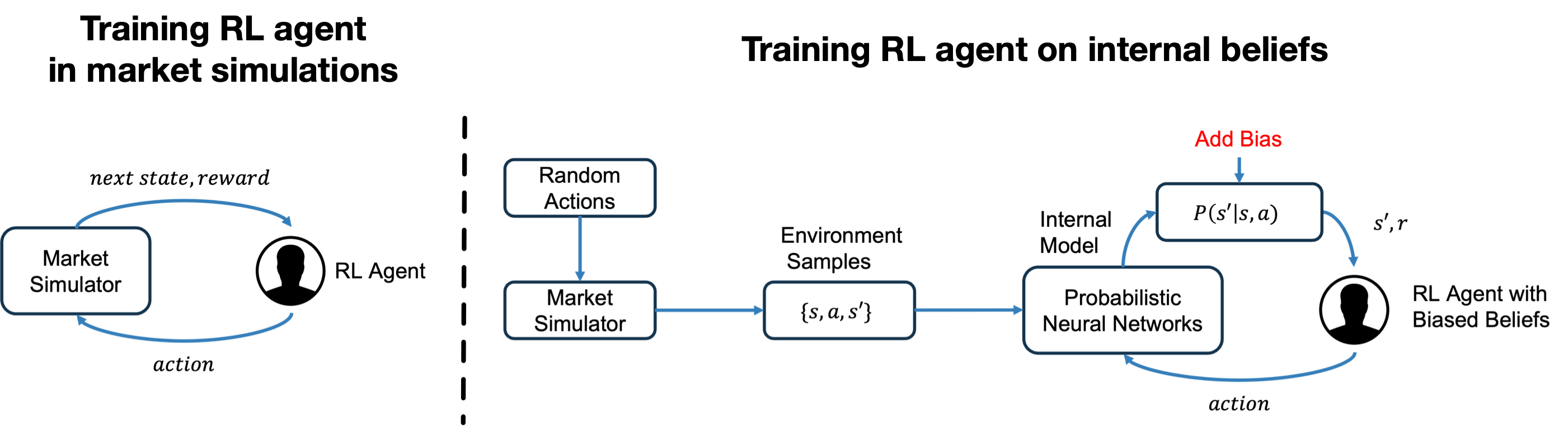}
    \caption{Training RL agents using market simulations or biased internal beliefs. (left) The RL agent learns a trading strategy by directly interacting with the simulated markets. (right) The RL agent learns from the internal model. We first learn a probabilistic internal model from the samples of the environment. We can inject bias to the internal model and use it to train a biased human.}
    \label{fig:internal_model}
    \vspace{-1ex}
\end{figure*}

To model the internal beliefs of a human trader, we collect samples of the environment by feeding random actions to the market simulator as illustrated in \cref{fig:internal_model}.
Each sample $(s, a, s')$ represents the transition of the environment state from $s$ to $s'$ when action $a$ is taken by the agent. 
% Each sample $(s, a, s', r)$ represents that an action $a$ is taken by the agent at time $t$, which changes the state from $s$ to $s'$ and gives a reward of $r$. 
We use the environment samples to train the internal model, which predicts the probability distribution of the next states given the current state and an action: $P(s'| s, a)$. In particular, we use a bootstrap ensemble of probabilistic neural network (\cite{janner2019trust}) which outputs the prediction as a parameterized Gaussian distribution with diagonal covariance $P(s'| s, a) = \mathcal{N} (\mu (s, a), \Sigma (s, a))$.
We validate the prediction results of the internal model in \cref{sec:internal_validation}.

Once we obtain the internal model, it can be used as a substitute of the real environment to provide the transition probability $P(s'|s,a)$ for training RL agents, while the rewards are given by the deterministic reward function in \cref{eq:reward}.
Adding bias to the internal model which is used in RL training will yield a biased human policy that is optimal with respect to the biased internal model, but sub-optimal in the real environment. 
% Adding bias to the outputs of the internal model will yield a biased human policy that is optimal with respect to the biased internal model, but sub-optimal in the real environment. 
For example, we can bias the internal model to overestimate the probability of positive $s'$ and use it to train an optimistic human agent. Similarly, we can modify the transition probability $P(s'|s,a)$ using \cref{eq:prospect_prob} to obtain a prospect biased human policy.
% The internal model can substitute the real environment by decomposing $p(s_{t+1}, r | s_t, a_t)$ into the transition probability $P(s'|s,a)$ and reward $R(s,a,s')$. we add bias to the model outputs and use it as the substitute of the real market to train biased human agents.

\section{Models of Human Sub-Rationality}\label{sec:model}
% A fully rational agent learns a optimal policy with high $\gamma$, that gives the optimal actions for ever state $s \in S$:
% \begin{equation}\label{policy}
%     \pi (s) = \text{argmax}_a \underset{s' \in S}{\sum}{P(s'|s,a)(R(s,a,s') + \gamma V(s'))}
% \end{equation}
While the Bellman equation (\cref{eq:bellman}) provides a solution to the optimal policy, it fails to model the deficiencies in human decisions. 
Based on their deviations from the optimal actions and the underlying driving factors, we classify human sub-rational behavior into two categories: (1) The \textbf{bounded (limited) human behavior} which show as random errors in decision-making, and (2) the \textbf{biased human behavior} that are systematically deviated from the optimal behavior.
% Inspired by the framework provided by \cite{Human_irrationality21}, we alter the $\gamma$ in \cref{eq:rational_policy} to model sub-rational human behavior caused by psychological bias, and modify the $\arg \max$ operator to model that caused by computational limitation.
% Inspired by the framework provided by \cite{Human_irrationality21}, we model the sub-rational human decision-making process by {\color{blue}SPECIFY ALTERING IN WHICH WAY - i.e. modifying which parameters}altering the above policy function \cref{eq:rational_policy}, which we use to study sub-rational trading behaviors.

\subsection{Bounded Human Behavior}\label{sec:bounded_rational}
To arrive at the optimal decision, one ideally needs comprehensive knowledge about all available choices and the ability to calculate the potential benefits. However, this assumption proves too challenging for humans in daily life due to their limited information access and computational power (\cite{simon1955behavioral}). Consequently, human decision-making tends to exhibit a certain degree of noise, and their decisions often incorporate random errors deviating from the optimal action.
\cite{simon1997models, simon1990bounded} proposed the idea of bounded rationality, which suggests that human are limited and tend to select a satisfactory decisions (approximately optimal) rather than the optimal decisions.
% This notion was first introduced by \cite{simon1997models, simon1990bounded}, where it is argued that human decision-making departs from perfect economic rationality. A perfectly rational decision requires access to information about all alternative choices, and the calculation of potential benefits. Since real humans typically do not have unlimited access to information and processing power, they are inclined to find satisfactory solutions to problems rather than the optimal solutions.

\begin{figure}[t]
    \centering
    \includegraphics[width=0.6\linewidth]{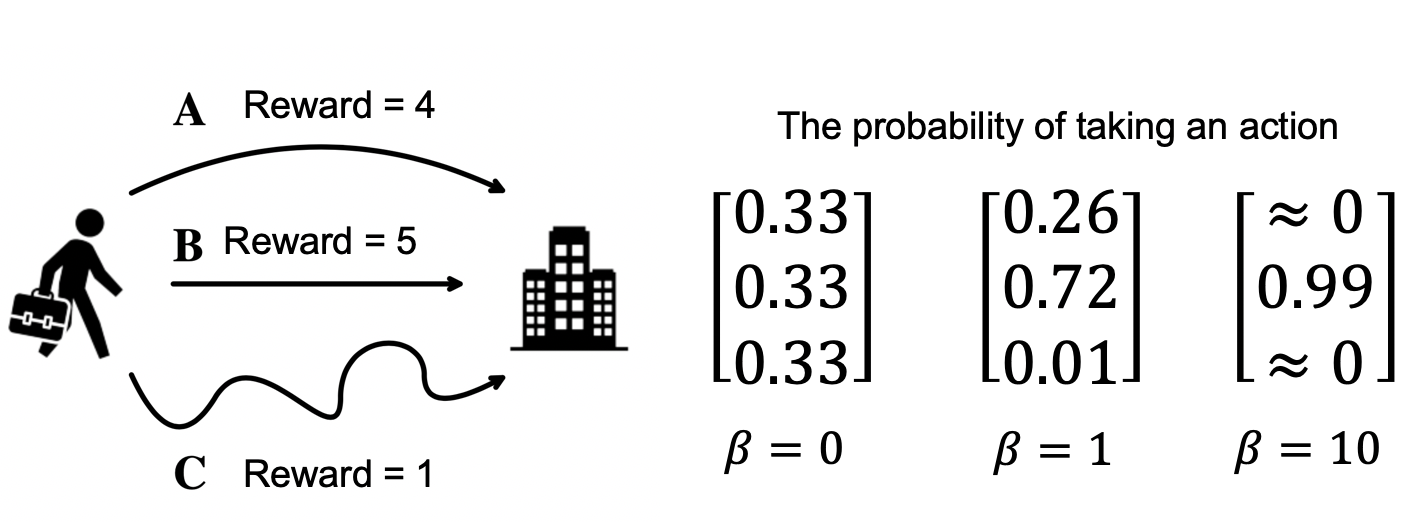}
    \caption{An example of the Boltzmann rationality model from \cite{laidlaw2022boltzmann} for three actions with different rewards (\textit{left}). For demonstration purpose here we simply consider a one step decision problem in a deterministic environment. The Boltzmann model gives the probability of taking an action using the $\beta$ parameter that adjusts the degree of rationality (\textit{right}). If $\beta = 0$, each action has the same probability to be selected. When $\beta = 10$, the model becomes more rational and only the action with highest reward is likely to be selected.}
    \label{fig:boltz}
    \vspace{-2ex}
\end{figure}

\begin{figure}[t]
    \centering
    \includegraphics[width=0.7\linewidth]{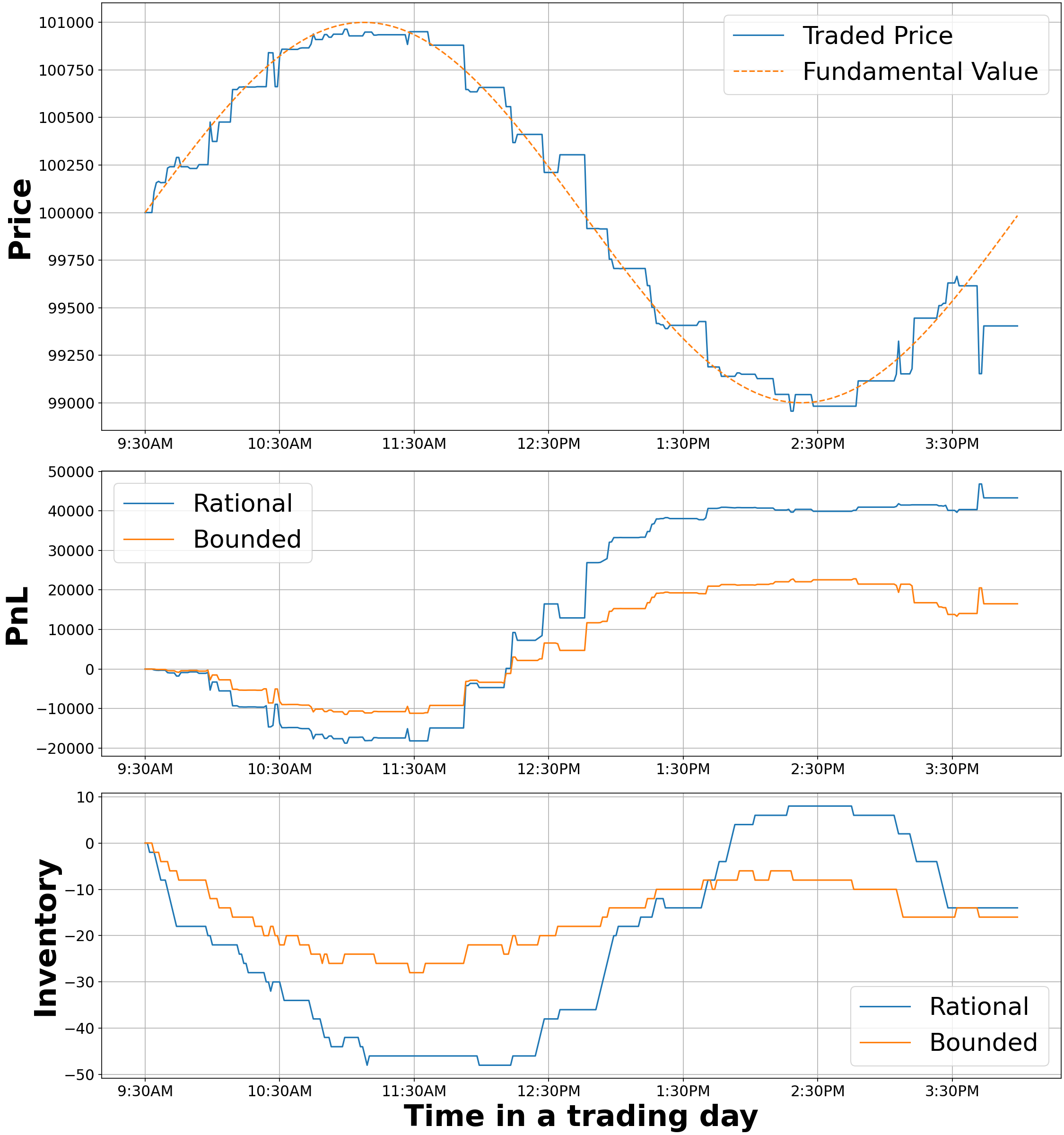}
    \caption{The behavior of bounded rational investors in the simulated market. Compared to the fully rational  investor, a bounded rational human investor takes sub-optimal actions that are similar but inferior. 
    % \penghang{To-do: replace the label of the agents to rational and sub-rational}
    }
    \label{fig:boltz_handcraft}
    \vspace{-2ex}
\end{figure}

\subsubsection{Boltzmann Rationality}
The most common approach for modeling bounded rational human decision-making is the Boltzmann rationality model (\cite{baker2005bayesian, ziebart2010modeling, asadi2017alternative}).
Unlike the rational policy in \cref{eq:rational_policy} that gives the optimal action using $\arg\max$, this model considers a probabilistic policy given by the Boltzmann softmax operator as follows:
% \textcolor{blue}{It is modeled by using a \textit{soft} approximation to the $\arg\max$~\cite{wikisoftmax} in (\ref{eq:rational_policy}) as follows:}
% Instead of using the argmax in~\cref{policy} to select the optimal action, the agent takes a probabilistic policy:
\begin{equation}
    \pi (a|s) = \frac{e^{\beta Q(s, a)}}{\sum_{a' \in A} e^{\beta Q(s, a')}}
    \label{eq:boltzmann}
    % \pi (a|s) = \text{Boltz}^{\beta}_a \underset{s' \in S}{\sum}{P(s'|s,a)(R(s,a,s') + \gamma V(s'))}
\end{equation}
where $Q(s, a) = \underset{s' \in S}{\sum}{P(s'|s,a)\left(R(s,a,s') + \gamma V(s')\right)}$ (\cite{asadi2017alternative}).
% where $\text{Boltz}^{\beta}_(x) = \frac{\sum_i x_i e^{\beta x_i}}{\sum_i e^{\beta x_i}}$~\cite{asadi2017alternative} \textcolor{blue}{<- This does not look like a probability, is this correct?}.
The Boltzmann softmax operator introduces a ``soft" optimization principle, which relaxes the assumption of unlimited resources and processing power, and allows the agent to take sub-optimal decisions with a preference for high-utility actions.
Figure \ref{fig:boltz} gives an example of modeling human behavior with Boltzmann rationality.
The parameter $\beta$ controls how likely the agent is to select the optimal action. 
If $\beta = 0$, the agent has zero intelligence and makes uniformly random decisions using the same probability to select every action.
As $\beta$ increases, the agent becomes more intelligent and makes less error in selecting the optimal action.
When $\beta \to \infty$, the Boltzmann operator approaches the argmax in~\cref{eq:rational_policy}, and the agent makes fully rational decisions. 

% \penghang{If you feel that the above explanation is sufficient to convey idea we can move the example to \cref{sec:handcraft} with the SHAP analysis.}

\noindent\textbf{An example of bounded rational trading behavior}
% \kshama{Do we want to say how we get the Q function to compute the bounded rational policy?}
\noindent To illustrate the bounded rational human trading behavior, we simulate a market with fundamental price following a sine wave and deploy a bounded rational human investor with $\beta = 0.2$ and a fully rational investor with $\beta = \infty$ (see \cref{eq:boltzmann}). The policy of the bounded rational human investor is obtained by passing the Q values of the fully rational investor through the Boltzmann softmax operator as in \cref{eq:boltzmann}. 
\cref{fig:boltz_handcraft} shows the market price and the actions of the two agents through a trading day. Since the market is simulated over a single time period of the sine wave with the same starting and closing price, the optimal strategy is to sell when the price is higher than the closing price (first half of the day), and  buy when the price is lower than the closing price (second half of the day).

Overall, we observe similar decisions of rational and bounded rational investors based on the holding position in \cref{fig:boltz_handcraft}: both of them tend to sell during the first half of the day, and buy during the second half of the day. 
However, the bounded rational human investor does not sell and buy at the maximum capacity compared to the fully rational investor that has a deeper holding position.
As a result, the bounded investor can not achieve the maximum reward even though the decisions are similar to those of the fully rational investor to some extent.
% In addition, we plot the distance from the limit order prices to the mid price in \cref{fig:boltz_handcraft}. 
% In addition, we plot the price of limit orders placed by both investors relative to the mid price in \cref{fig:boltz_handcraft}.
% Compared to the bounded rational human investor, the electronic investor is more confident and often places the order deeper in the LOB. Limit orders placed far from the mid price yield more rewards as the agent is buying and selling at a better price, but have higher risk of not being executed.

In summary, the bounded rational model emulates human traders with limited information availability and processing capacity. We observe that such limitations lead to sub-optimal decision-making that is similar albeit inferior compared to the homo economicus. 

\subsection{Biased Human Behavior}
Here we describe several human cognitive and psychological biases, which lead to systematic deficiencies in human decisions.

\subsubsection{Myopic Behavior} \label{sec:myopia}
Humans in reality may be psychologically biased to only care about the short-term results. They make myopic decisions without considering how the actions affect them far into the future.
% investors demonstrate myopic behavior when they care only about short-term results with absolutely no consideration of how a certain action may affect them in the future.
% Myopic behavior is a behavior based on the pursuance of short-term results and represents an action in regards to what one wants now, without taking into account any future consequences.
% The sense behind the financial myopia illusion is that investors can use current news and other market data available online to understand which adjustments need to be made to increase the surety of achieving the investors’ goals.
% One can become too anxious and distracted by constantly evaluating short-term market performance. For example, it can incentivize a deliberate and thoughtful investor to abandon the existing long-term-oriented investment plan.
A typical example of myopic human behavior in finance is myopic loss aversion~(\cite{thaler1997effect, benartzi1999risk, chrisman2012variations}). Investors that focus on short-term return may react too passively to recent losses. As a result, they abandon their existing long-term-oriented investment plan and lose the potential to achieve better benefits in the future. 
Studies have shown that financial media can often facilitate myopic trading behavior by constantly reporting market news and portraying a sense of urgency to act. Investors who receive such information too frequently tend to avoid investing in riskier assets that may yield better long-term rewards~(\cite{larson2016can}).

\begin{figure}[t]
    \centering
    \includegraphics[width=0.5\linewidth]{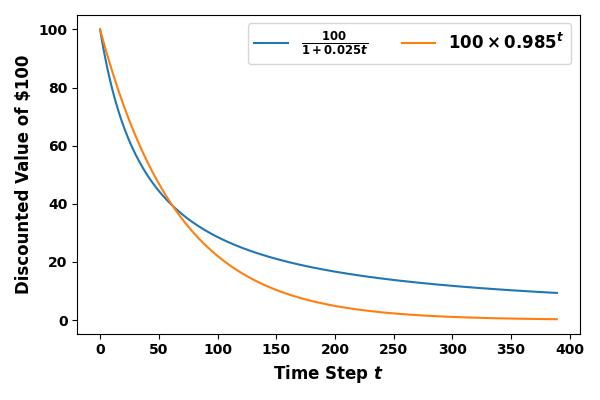}
    \caption{Comparison of exponential discounting to hyperbolic discounting of \$100 over $390$ time steps. See the steady drop in discount rate for the \textcolor{orange}{orange} exponential curve. On the other hand, the \textcolor{RoyalBlue}{blue} hyperbolic curve shows steep discounting early on and modest discounting later on.}
    \label{fig:hyp_vs_exp}
\end{figure}

Preferences of humans when choosing between short-term and long-term rewards is modeled by discounting the long-term rewards, and have been studied in literature on economics and psychology (\cite{chabris2010intertemporal,grune2015models}). Two popular ways to model this temporal discounting of rewards include the exponential discounting model (\cite{samuelson1937note}) and the hyperbolic discounting model (\cite{ainslie1992picoeconomics}). Numerically, the exponential discounting model expresses the present value of a reward $r$ given $t$ time steps from now as \begin{align}
    V=r\gamma^t
    \label{eq:exponential}
\end{align}
where $\gamma\in[0,1]$ is the exponential discounting factor. In contrast, the hyperbolic discounting model expresses the present value of the reward $r$ given with delay $t$ as \begin{align}
    V=\frac{r}{1+kt}
    \label{eq:hyperbolic}
\end{align}
where $k\geq0$ is the hyperbolic discounting factor. \cref{fig:hyp_vs_exp} shows a comparison of both temporal discounting schemes for median values of discounting parameters observed in real humans in \cite{green1997rate}. For simplicity and amenability to standard RL algorithms, we assume that the discount factors $\gamma$ and $k$ do not depend on the amount of reward $r$ although there have been empirical studies that demonstrate that the rate of discounting reduces with an increase in the amount of reward (\cite{green1997rate}).

Empirical evidence from humans and pigeons support the hyperbolic discounting model as it captures reversals in preference between delayed rewards (\cite{green1994temporal,mazur1985probability}). This is described in the following scenario. Suppose one is given a choice between receiving \$$10,000$ today versus receiving \$$11,000$ in a week. Most people would choose the former option to receive \$$10,000$ today considering the unknown factors that could prevent the receipt of the delayed \$$11,000$. Now suppose the choice was between receiving \$$10,000$ in a year versus receiving \$$11,000$ in a year and one week. In this scenario with rewards delayed further, most people would prefer to wait the extra week to receive \$$11,000$. This is based on the reasoning that the chances of not receiving \$$11,000$ after a year and a week are lower conditioned on having received \$$10,000$ after a year. This change in preference between the smaller and larger rewards when the delay to both is increased is called preference reversal. Exponential discounting does not capture preference reversal since the discount rate remains the same for all values of delay. On the other hand, the discount rate in the hyperbolic discounting scheme is high in the short-term and slows down in the long-term as can be seen in \cref{fig:hyp_vs_exp}.

While it maybe more pragmatic to consider hyperbolic discounting schemes for our myopic human model, non-exponential discounting poses computational challenges to standard RL algorithms that are based on the validity of the Bellman equation. There is work on formulating temporal difference learning schemes for hyperbolic discounting (\cite{alexander2010hyperbolically,redish2010neural})
% \cite{alexander2010hyperbolically} propose a TD learning method for hyperbolic discounting that was later shown to not converge to the hyperbolic value function \textcolor{red}{add citation!}. 
\cite{redish2010neural} and later \cite{fedus2019hyperbolic} propose approximating the policy for hyperbolic discounting using a combination of many exponentially discounting policies. In particular, \cite{fedus2019hyperbolic} propose a solution to approximate the hyperbolic Q function using a weighted sum of exponential Q functions that are each computed for a different exponential discount factor. This work allows for the use of deep neural networks to approximate the underlying exponential Q functions. 
This adds a computational overhead of learning many RL policies that use different exponential discount factors alongside the instability issues that arise with the use of off-policy RL algorithms such as Deep Q-Learning that estimate the Q function (\cite{mnih2013playing,haarnoja2018soft}).
% A disadvantage of this approach being that it is not directly applicable to popular policy-gradient RL algorithms such as PPO that are typically more stable since they use a state value function \cite{maei2009convergent}. 
%Previous work based on \cite{redish2010neural} on learning many exponentially discounting $\mu$agents and combining them to get a hyperbolically discounting agent.
Therefore, for the purpose of this paper, we consider exponential discounting as a model for myopicity of human traders.

Recall that the discount factor $\gamma$ in equation \cref{eq:bellman} models decay in the value of rewards with time delay (\cite{chabris2010intertemporal}).
To model myopic human investors, we decrease the $\gamma$ in the Bellman equation \cref{eq:bellman} and corresponding policy \cref{eq:rational_policy}. When $\gamma \approx 1$, the agent is fully rational and considers both short-term and long-term rewards. As $\gamma \to 0$, the agent becomes myopic and only acts to maximize the immediate reward $R(s,a,s')$.

\begin{figure}[t]
    \centering
    \includegraphics[width=0.8\linewidth]{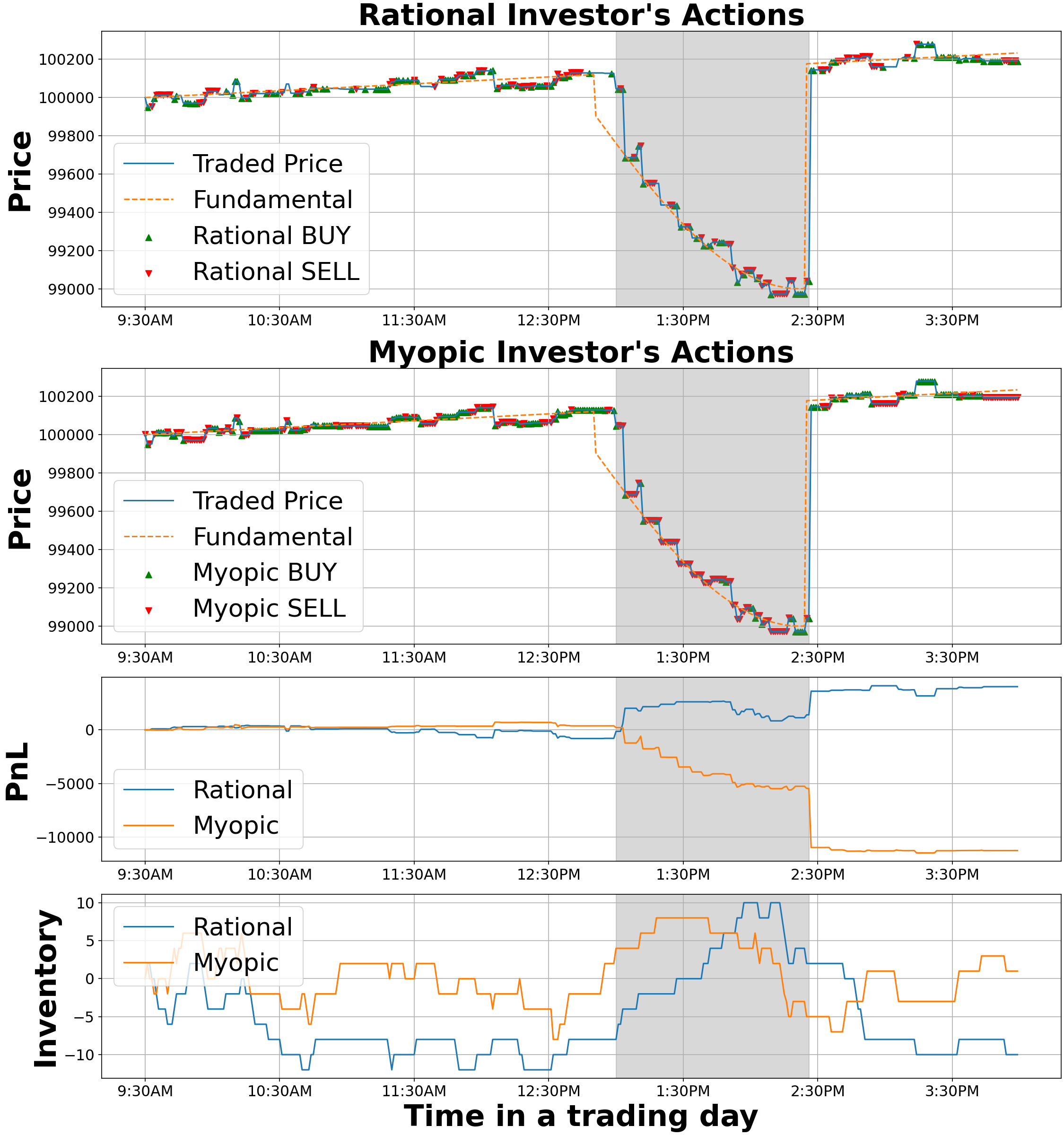}
    \caption{The behavior of myopic investors in the simulated market. Compared to the fully rational investor, a myopic human investor reacts too passively to the short-term losses and abandons its long-term investment plan.
    % \penghang{Myopic human has lower holding, connections between myopic loss aversion, prospect bias, and inventory risk.}
    }
    \label{fig:myopic_handcraft}
    \vspace{-3ex}
\end{figure}

\noindent\textbf{An example of myopic trading behavior}

\noindent Here we give an example to distinguish the behavior of myopic investors from that of fully rational investors, using a hand-crafted market simulation. The human investor is trained in the default market simulations with Ornstein-Uhlenbeck fundamental with $\gamma = 0.1$, while the rational investor is trained with $\gamma = 0.99$ (see \cref{eq:rational_policy}).
The trained investors are then tested in a hand-crafted market scenario with a different fundamental price as described here.
The trading day can be decomposed into three stages. In the first and last stages, the price increases with a linear trend, while the market is under a shock in the second stage with the price temporarily dropping by 1\%. \cref{fig:myopic_handcraft} shows the market prices and the actions of the agents through the trading day. \cref{tab:MLA} displays the ratio of buy orders to sell orders for both investors, alongside the percentage of hold decisions (do not buy or sell) relative to all decisions in each of the three stages of the market.
% We simulate the market in three different stages. 
Since the asset price increases continuously in stage 1, both rational and myopic investors place more buys than sells (\cref{tab:MLA}), and achieve similar rewards.
When the market is under a shock in stage 2, the myopic human investor tries to sell out their shares to mitigate the loss from the temporary price drop. 
They place more sell orders than buy orders, and their inventory drops significantly during the second stage. The rational investor, on the other hand, shows better vision of the future and is less affected by the shock. They place similar amount of buy and sell orders, and their inventory increases as the buy orders are easier to be executed during the shock in stage 2.
By the end of the day, the rational investor obtains a significantly better PnL as they bought a lot of shares at low prices, compared to the myopic human investor who sold a lot of shares during the shock. 

Overall, the myopic investor demonstrates a short-term trading strategy with a much smaller tendency to hold (and not trade) as compared to the rational investor especially during the shock (\cref{tab:MLA}).
% Overall, the myopic investor demonstrates a short-term trading strategy wherein they do not tend to hold like the rational investor does in more than 34\% of the time (\cref{tab:MLA}). 
Subsequently, their inventory fluctuates between long and short positions more frequently than that of the rational investor (\cref{fig:myopic_handcraft}).
In summary, the hand-crafted market scenario demonstrates that our model of psychologically myopic behavior successfully reproduces the myopic investing behavior observed in financial markets, i.e., myopic loss aversion (as described in \cref{sec:myopia}).

\subsubsection{Prospect Theory}\label{sec:prospect}
A well-known model of decision making under uncertainty is the expected utility theory, which measures the utility of the outcomes as:
\begin{equation}\label{eq:expected_utility}
    V = \sum^n_{i=1} p_i \cdot v(x_i)
\end{equation}
where $v(x_1), v(x_2), \dots, v(x_n)$ are the values of the potential outcomes and $p_1, p_2, \dots, p_n$ are their respective probabilities. The objective function of an RL agent is in line with the expected utility theory, as in \cref{eq:bellman} the value of a state is measured by the sum of utilities of the potential next states weighted by their transition probabilities.

\begin{table}[t]
    \centering
    \caption{The decisions of investors during the three stages in \cref{fig:myopic_handcraft}.}
    \makebox[\textwidth]{
    \begin{tabular}{|l|c|c|c|c|c|c|}
    \hline
         \multirow{2}{*}{}& \multicolumn{3}{c|}{Buy / Sell Ratio} & \multicolumn{3}{c|}{Hold}\\ \cline{2-7}
         & Stage 1 & Stage 2 (shock) & Stage 3 & Stage 1 & Stage 2 (shock) & Stage 3 \\ \hline
         Rational & 1.38 & 0.94 & 1.33 & 38\% & 34\% & 38\% \\ \hline
         Myopic & 1.49 & 0.41 & 1.26 & 2\% & 1\% & 7\% \\ \hline
    \end{tabular}
    \label{tab:MLA}
    }
\end{table}

However, a substantial body of evidence has shown that human decision makers systematically violate the expected utility theory. To model such deviation, Kahneman and Tversky introduced prospect theory (\cite{kahneman1979prospect}) which consists of two key elements: 
\begin{itemize}
    \item A value function that is concave for gains, convex for losses, and steeper for losses than gains.
    \item A nonlinear transformation of the probability scale, which overweights small probabilities and underweights moderate and high probabilities.
\end{itemize}

\noindent\textbf{Prospect Biased Utility}

\begin{figure*}[t]
    \includegraphics[width=0.44\linewidth]{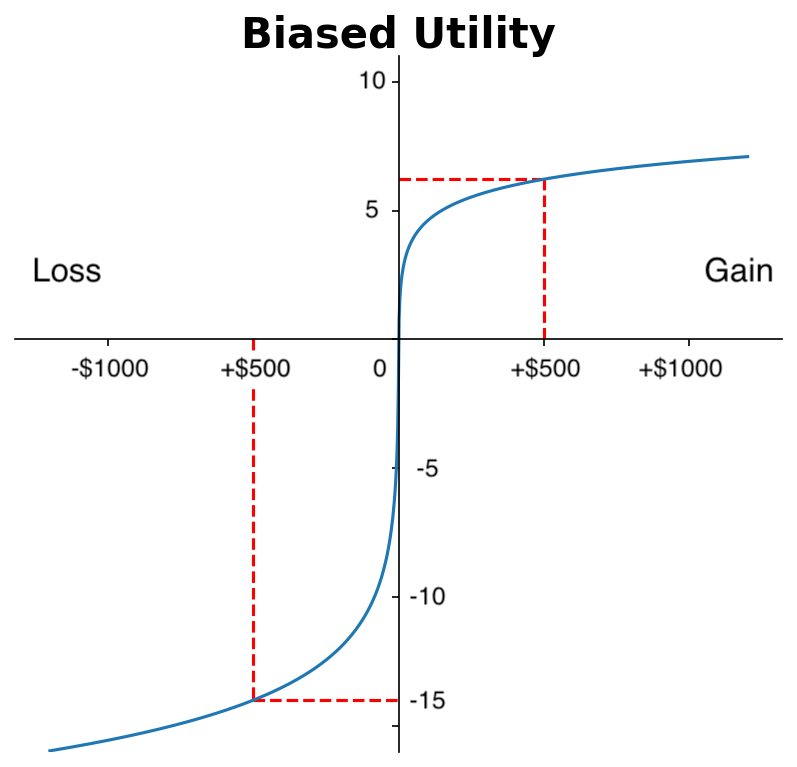}
    \hspace{1ex}
    \includegraphics[width=0.44\linewidth]{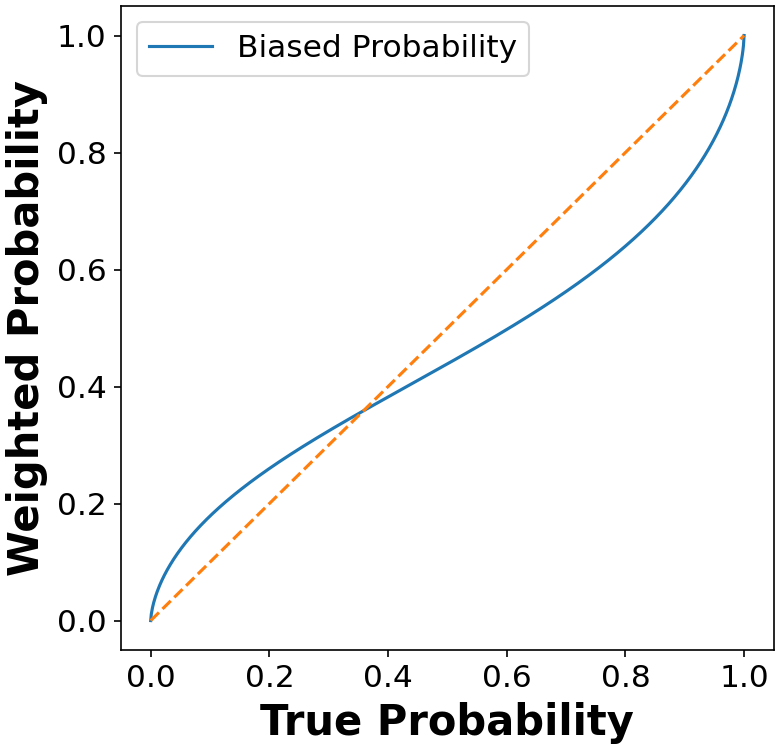}
    \caption{(\textit{left}) The prospect biased utility function. The x-axis represents the amount of gain/loss and the y-axis is the corresponding weighted value. (\textit{right}) The prospect biased probability function. The x-axis is the true probability and the y-axis shows the weighted probability.}
    \label{fig:prospect_theory}
    \vspace{-1ex}
\end{figure*}
\noindent The first element of prospect theory stems from loss aversion, which refers to the fact that humans feel losses more than two times greater than the equivalent gains. Empirical studies (\cite{kahneman1979prospect}) have shown that human's value function is concave for gains and convex for losses. 
For example, human prefer to choose 100\% chance to win \$490 rather than 50\% chance to win \$1000, even though the expected utility of the latter is higher. As shown in \cref{fig:prospect_theory} (left), the prospect biased utility of \$490 gain is larger than a half of the \$1000 gain. Therefore the certain gain is more preferable as $v(+\$490) > 0.5 v(+\$1000) \rightarrow 100\% \times v(+\$490) > 50\% \times v(+\$1000)$.
On the other hand, humans prefer to choose a 50\% chance to avoid a \$1000 loss rather than accepting a certain loss of \$490 with higher expected utility. As shown in \cref{fig:prospect_theory} (left), the prospect biased utility of a \$490 loss is lower than a half of the \$1000 loss, i.e., $v(-\$490) < 0.5 v(-\$1000) \rightarrow 100\% \times v(-\$490) < 50\% \times v(-\$1000)$.

Note that loss aversion results in both risk averse behavior in gains and risk seeking behavior in losses. This differs from the risk-sensitive RL approaches (\cite{mihatsch2002risk, vyetrenko2019risk}) that set a risk preference parameter for the agent to be either risk-averse, risk-neutral, or risk-seeking, as well as other economic analyses that commonly only assume risk aversion.

Inspired by \cite{Human_irrationality21}, we modify the reward function in the Bellman equation to $V(s) = \underset{a}{\max} \underset{s' \in S}{\sum}{P(s'|s,a)\left(f_c(R(s,a,s')) + \gamma V(s')\right)}$, where
\begin{equation}
    f_c(r) =
    \begin{cases}
      \log (1 + |r|) & \text{if $r > 0$}\\
      0 & \text{if $r=0$}\\
      -c \log(1 + |r|) & \text{if $r < 0$}
    \end{cases}. 
\label{eq:prospect_utility}
\end{equation}
In the equation, $c$ is the coefficient of loss aversion ($c > 1$ as human naturally feel losses greater than gains). While the agent is still risk-seeking in losses and risk-averse in gains regardless of the coefficient, an agent with larger $c$ value will be more risk averse when comparing the same amount of potential gains and losses.

\noindent\textbf{Prospect Biased Probability}

\noindent Another aspect of prospect theory is that humans do not perceive probability in a linear fashion. When addressing any uncertainty, humans consider the natural boundaries of probability, impossibility and certainty, as the two reference points. The impact of a change in probability diminishes with an increase in its distance to the reference points. Considering a 0.1 increase in the probability of winning a prize, a change from 0 to 0.1 chance to win has more impact than a change from 0.45 to 0.55. To model this distortion in probability seen in humans, \cite{tversky1992advances} introduce a non-linear weighting function as follows,
\begin{equation}\label{eq:prospect_prob}
    w(p) = \frac{p^\delta}{(p^\delta + (1 - p)^\delta)^{1/\delta}}
\end{equation}
where $p$ is the probability of a potential outcome, and $\delta$ is estimated to be 0.61 and 0.69 for gains and loses respectively (here for simplicity we set $\delta = 0.65$ for all types of outcomes). As shown in \cref{fig:prospect_theory} (right), humans overweight events with low probabilities and underweight events with high probabilities. We apply the above weighting function in the Bellman equation $V(s) = \underset{a}{\max} \underset{s' \in S}{\sum}{w(P(s'|s,a))\left(R(s,a,s') + \gamma V(s')\right)}$ to model prospect biased probability in the RL setting.

\begin{figure}[t]
    \centering
    \includegraphics[width=0.8\linewidth]{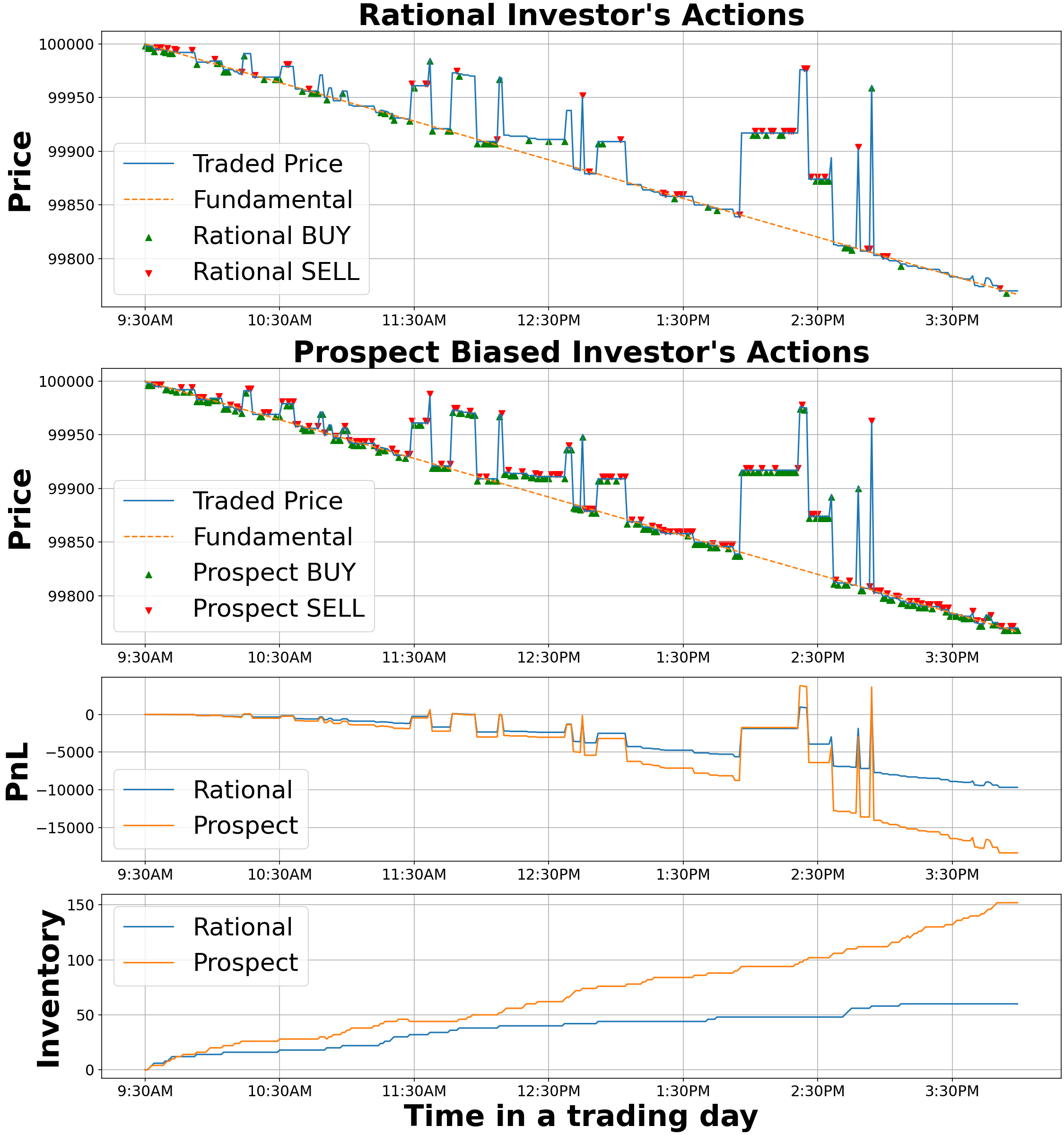}
    \caption{The behavior of prospect biased investors ($c=2.5$ and $\delta=0.65$) in the simulated market. When experiencing losses (negative PnL, see period around 2pm), the rational investor decides to hold more often and does not increase its inventory. In contrast, the prospect biased investor is more willing to take risks to avoid losses. They buy more shares with the hope of reducing their average purchase price, which allows them to avoid losses when the stock price temporarily increases around 2:20pm.}
    \label{fig:prospect_handcraft}
    \vspace{-2ex}
\end{figure}

\noindent\textbf{An example of prospect biased trading behavior.}

\noindent\cref{fig:prospect_handcraft} demonstrates how prospect biased behavior deviates from the optimal trading strategy. The market price continuously decreases in the entire day, and both investor agents are not able to gain by the end of the day. Since they experiencing losses throughout the day, it allows us to investigate how rational and prospect biased investor handle losses differently. In the first half of the day, the rational and prospect biased investor have similar holding positions. After recognizing the long-term market trend, the rational investor stops trading frequently and does not increase the holding in the second half of the day. 
When the market becomes extremely volatile in a short time (between 1:50PM and 2:40PM), the prospect biased investor addresses the uncertainty with a risk-seeking strategy. Due to the convex utility function \cref{eq:prospect_utility} for losses and over-weighted low probabilities (\cref{eq:rational_policy}), they prefer to gamble on actions that can mitigate or avoid losses even if the expected utility is low and the success probability is small. As a result, they decide to buy more shares after the price drop, which reduces the average price at which the stock is purchased. This allows them to offset their losses once the stock return to the average purchased price. In fact, the prospect biased investor is able to achieve positive PnL during 2:20 pm. However, such strategy increases the overall risk exposure, and the PnL is highly volatile.

In summary, our model successfully capture human bias described in the prospect theory literature (\cite{kahneman1979prospect}). In the handcrafted market scenario, the prospect biased investor tend to be risk-seeking when facing choices between losses.

\subsubsection{Optimism/Pessimism}\label{sec:opti_pessi}
Humans may have biased expectations of the future. \cite{sharot2007neural} show that humans often systematically overestimate or underestimate their chances of experiencing positive and negative events. Inspired by \cite{Human_irrationality21}, we model the optimistic and pessimistic humans by modifying the Bellman equation to:
\begin{align}
    V(s) = \underset{a}{\max} \underset{s' \in S}{\sum}{P^\omega(s'|s,a)\left(R(s,a,s') + \gamma V(s')\right)}\label{eq:optimism}\\
    \textnormal{where }P^\omega(s'|s,a) = \frac{P(s'|s,a)e^{\omega\left(R(s,a,s') + \gamma V(s')\right)}}{\underset{s' \in S}{\sum}{P(s'|s,a)e^{\omega\left(R(s,a,s') + \gamma V(s')\right)}}}.
\end{align}
% where $P^\omega(s'|s,a) = P(s'|s,a)e^{\omega\left(R(s,a,s') + \gamma V(s')\right)}$. 
% \kshama{Is there a normalizing quantity omitted here?}
When $\omega = 0$, \cref{eq:optimism} considers the original transition probability and the agent is rational. 
% For $\omega\neq0$, the agent modifies the transition probability based on the value of the state.
As $\omega \to +\infty$ (or $\omega \to -\infty$), the agent becomes optimistic (pessimistic) by overestimating the probability of positive (negative) transitions.

% \penghang{To-do: tried a few market scenarios but so far none of them give a intuitive explanation of optimistic/pessimistic behavior. The first reason is that exponentially overweight/underweight the reward based on its values seems to be too extreme. An outlier of large reward will take over the probability mass. Another reason is that we do not prevent the cash/holding to go below zero so the agent learns a extremely aggressive policy, i.e., only buy (or sell) no matter what happens.}
% \kshama{Do you think normalizing the rewards to lie in maybe [-1,1], and using small range of $\omega$ values may help? We need to make sure we normalize correctly -- by fixed amount so that only scale varies (and other experiments still hold).}
\begin{figure}[t]
    \centering
    \includegraphics[width=0.48\linewidth]{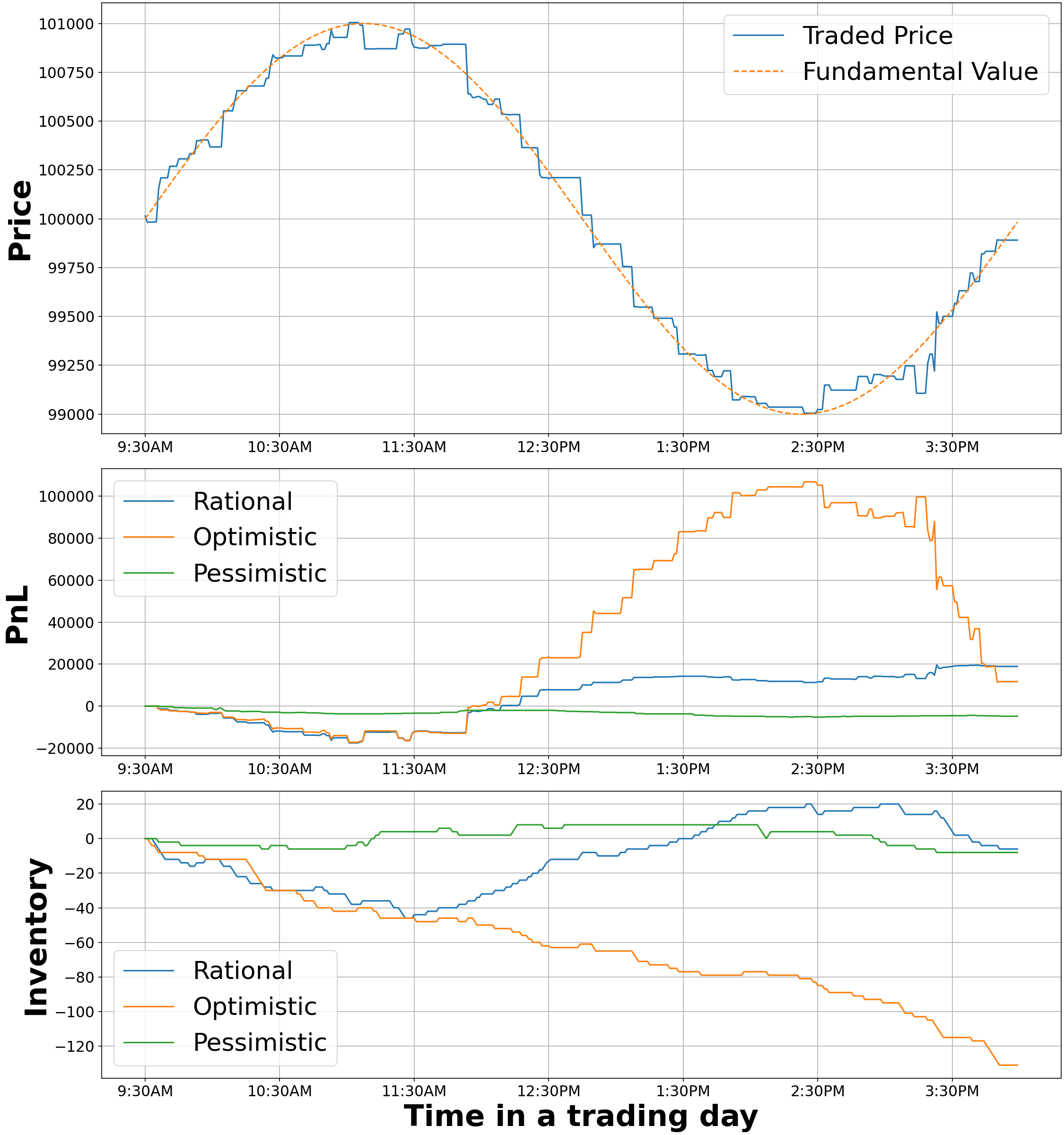}
    \includegraphics[width=0.48\linewidth]{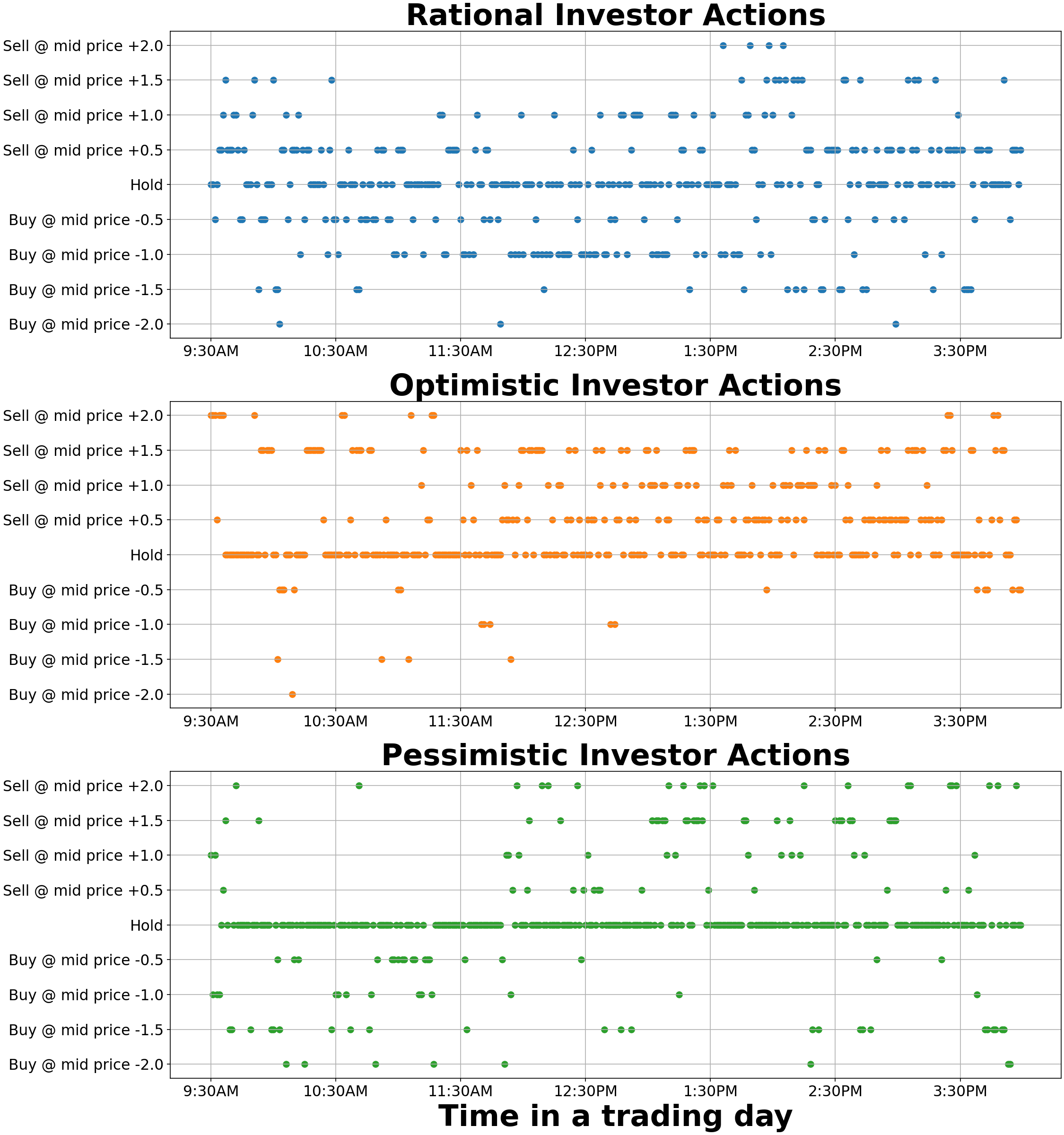}
    \caption{The behavior of optimistic and pessimistic investors ($\omega=1$ and $\omega=-1$ respectively) in the simulated market. (left) The market prices and agents' portfolio values and inventories. (right) The actions of the rational, optimistic, and pessimistic traders. Compared to the rational agent (blue), the optimistic agent (orange) tends to place orders aggressively on one side of the LOB, while the pessimistic (green) agent prefer to hold in most of the time.}
    \label{fig:opti_pessi}
    \vspace{-2ex}
\end{figure}

\noindent\textbf{An example of optimistic/pessimistic trading behavior}

\noindent We illustrate the behavior of optimistic and pessimistic investors in simulated markets by adopting a fundamental price which follows a sine wave as shown in \cref{fig:opti_pessi}. Similar to observations in \cref{fig:boltz_handcraft}, the rational investor takes a optimal strategy by selling at the higher prices (first half of the day) and buying at the lower prices (second half of the day). The optimistic agent has a more aggressive strategy compared to the fully rational investor. While they have similar holding positions between 9:30am and 12:00pm, the optimistic investor builds a strong belief that the market trend will remain the same and continues short selling even when the price is lower than the starting price. The portfolio value of the optimistic investor is extremely volatile due to their over-confident trading strategy: they obtain massive gains when they are right (from 12:00pm to 2:30pm) and massive losses when they are wrong (from 2:30pm to 4:00pm). On the other hand, the pessimistic investor trades very little as can be seen by the large number of hold decisions due to their under-confidence, and do not have any hope of profitable investments.
% On the other hand, the pessimistic investor is under-confident and does not have any hope of profitable investments. 
Additionally, the pessimistic investor has a small inventory and is less affected by the market trends. 
% {\color{blue}This is really nicely described!}

% \penghang{To-do: find some citations.}

\section{Results}\label{sec:results}
% \penghang{Work in progress.}
In this section, we examine the impact of sub-rationality of human strategies on:
% In this section, we examine the trading strategies of sub-rational human investors based on three key aspects: 
(1) their daily profit and loss (PnL) in simulated markets, 
(2) the importance given to various state features through an explainability analysis of their policy, and
% (2) the SHAP explanation of their policy, and 
(3) market observables.
% \begin{enumerate}
%     \item How does the degree of sub-rationality affect the agent's ability to achieve rewards?
%     \item What are the differences between sub-rational investor behavior and fully rational investor behavior?
%     \item How does sub-rational trading behavior affect the financial market?
% \end{enumerate}
We train and test the RL investors in markets simulated with ABIDES-gym using the same configuration described in \cref{sec:simulation}.
% We consider a default market environment with 2 value agents, 1 market maker, 2 momentum agents, and 20 noise agents. The fundamental value of the market is generated as a stochastic Ornstein-Uhlenbeck process. To address the first question, we add a human investor with different degree of sub-rationality to the default market scenario, and evaluate the profits during training and testing (\cref{sec:perfomance}). We then apply the human investor alongside a fully rational electronic investor in the market with the same background trading agent configuration, but with a hand-crafted fundamental series to address the second question. For the last question, we add $10 \lambda$ human investors and $10 (1-\lambda)$ electronic investors for $\lambda\in[0,1]$ to the default market and compare the market observables (\cref{sec:impact_of_human}).
\begin{figure*}[t]
    \centering
    \includegraphics[width=0.45\linewidth]{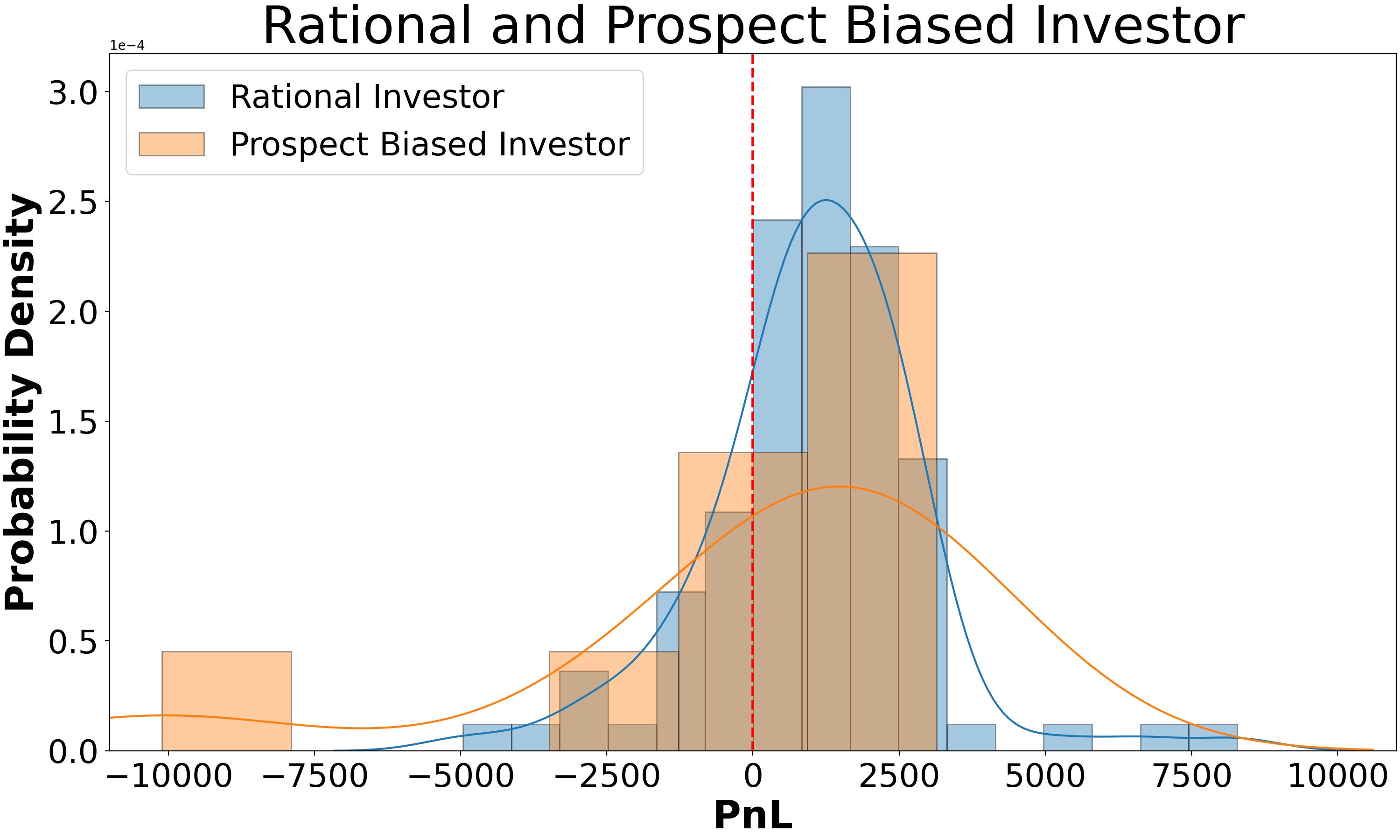}
    \includegraphics[width=0.45\linewidth]{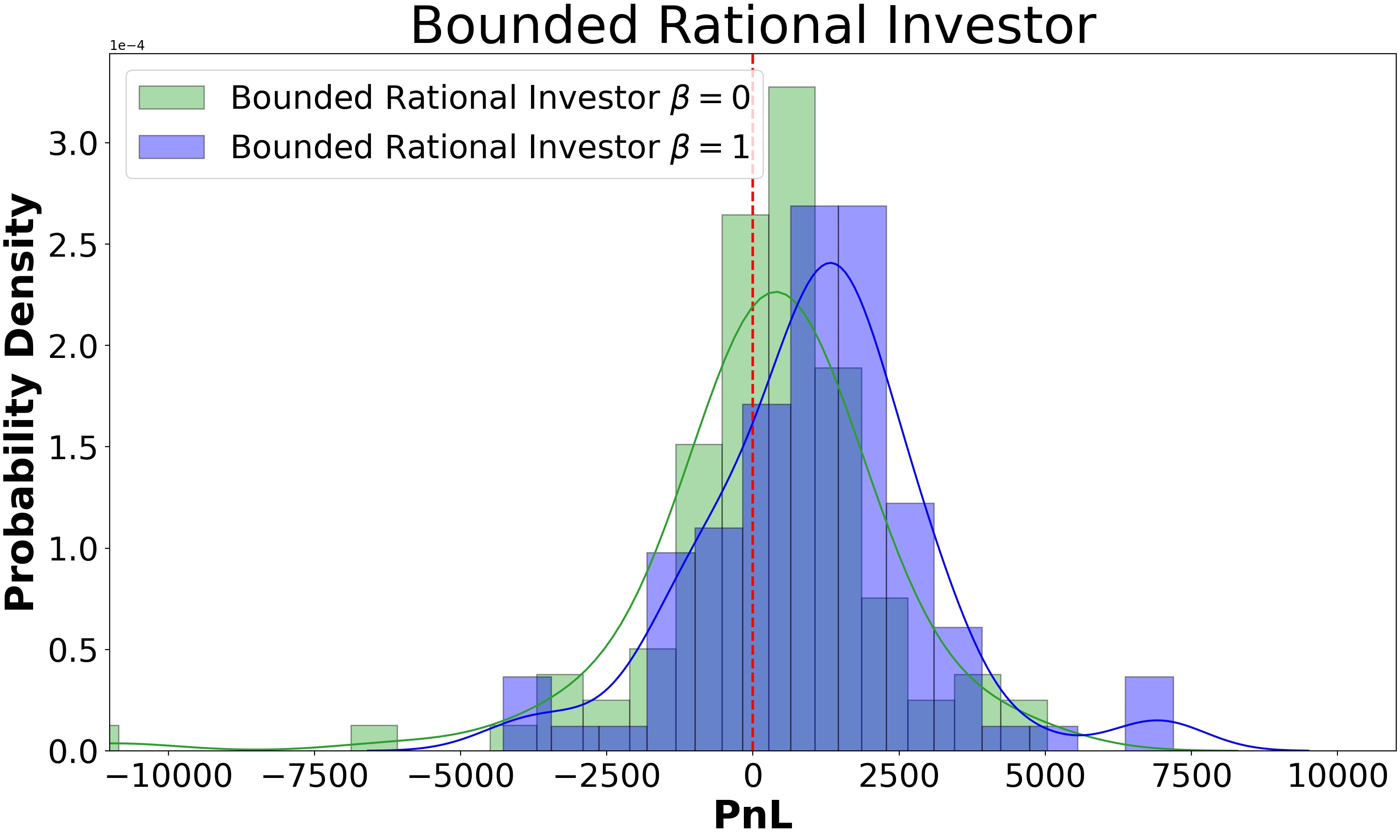}
    
    \includegraphics[width=0.45\linewidth]{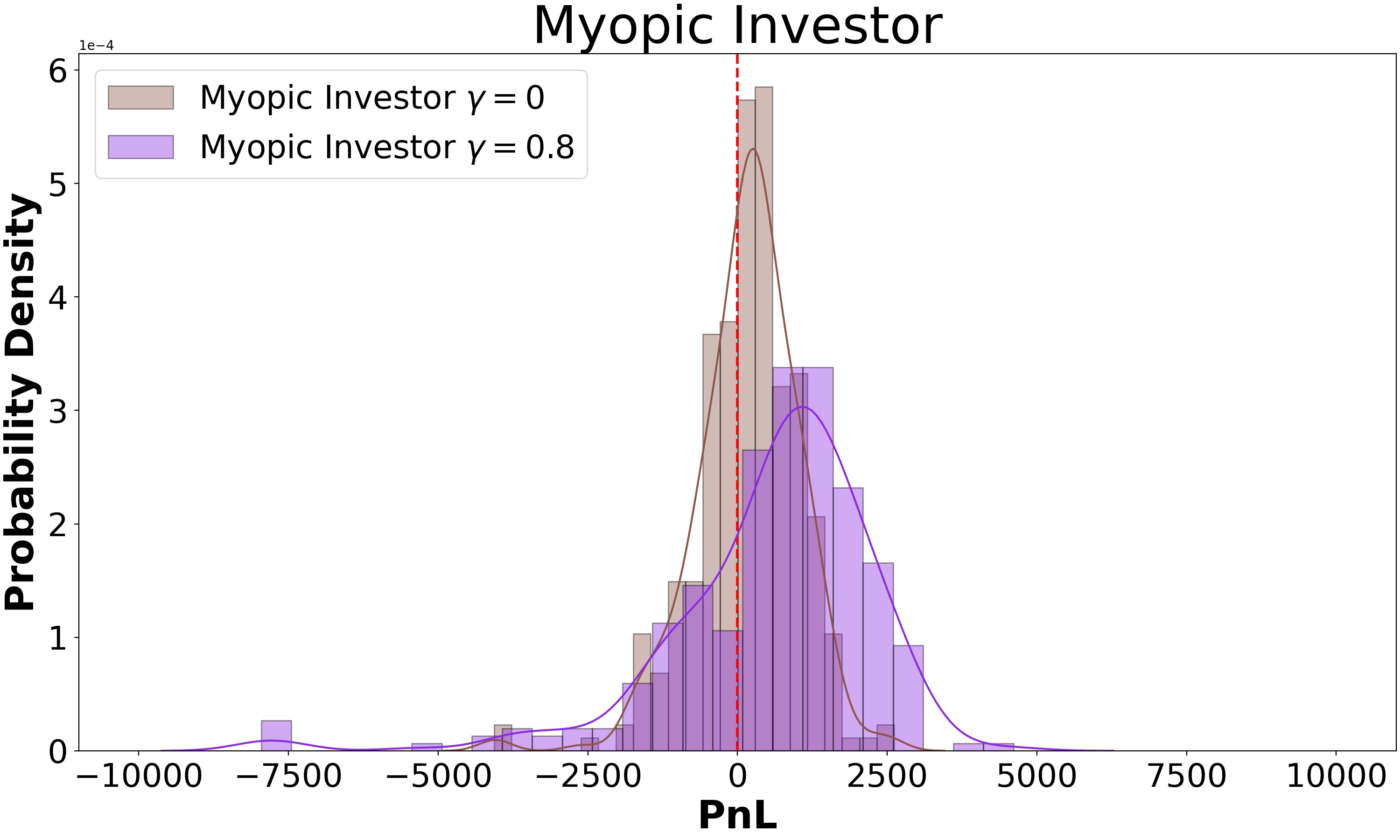}
    \includegraphics[width=0.45\linewidth]{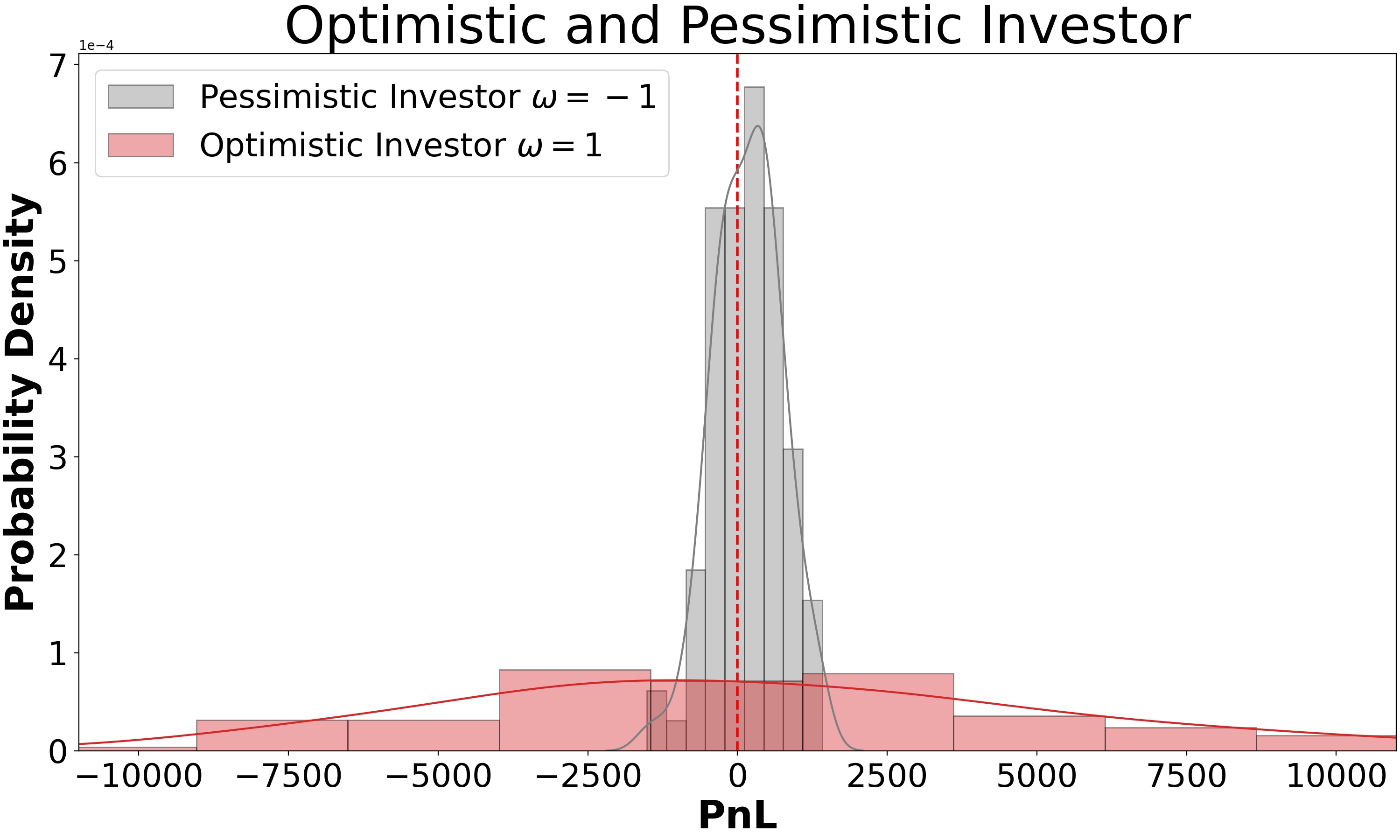}
    
    \caption{The PnL of sub-rational human investors in simulated markets. 
    % \kshama{Minor: Wondering if we want to have the same x limits for all plots, and add vertical lines for the means of the distributions?} 
    % \penghang{The x limits are same now. We don't have to plot the means as they are listed in \cref{tab:pnl}.}
    }
    \label{fig:pnl}
    \vspace{-1ex}
\end{figure*}

\begin{table}[b]
    \centering
    \caption{The mean and standard deviation of the PnL of sub-rational human investors.}
    \makebox[\textwidth]{
    \small
    \begin{tabular}{|l|c|c|c|c|c|c|c|c|}
    \hline
         \multirow{ 3}{*}{Metrics} & \multirow{ 3}{*}{Rational} & Prospect Biased & Bounded & Bounded & Myopic & Myopic & Pessimistic & Optimistic \\ 
          & & $c=2.5$  & (zero-intelligence) &  & (fully-myopic) &  & & \\
          & & $\delta=0.65$ & $\beta=0$ & $\beta=1$ & $\gamma=0$ & $\gamma=0.8$ & $\omega=-1$ & $\omega=1$ \\
         \hline
         Mean & 1060.42 & 117.3 & 29.75 & 1056.35 & 162.76 & 575.98 & 174.11 & 454.93 \\ \hline
         Std &  1835.58 & 3740.26 & 2711.04 & 1927.67 & 867.96 & 1732.98 & 561.98 & 5487.65 \\ \hline
    \end{tabular}
    }
    \label{tab:pnl}
\end{table}

\subsection{Profit and Loss of Sub-Rational Investors}\label{sec:perfomance}
We first assess the profit and loss (PnL) of the sub-rational human investors in the simulated markets. In particular, we test each type of sub-rational trading strategy over 100 simulated trading days and show the distribution of the single-day PnL in \cref{fig:pnl} and \cref{tab:pnl}.
Formally, we define the single-day PnL as the change in portfolio value by the end of a trading day (390 minutes in total from 9:30am to 4:00pm) in cents:
$$\text{PnL} = \text{PortfolioValue}_{end} - \text{PortfolioValue}_{start}$$

We observe that the rational investor has the best overall performance in simulated markets. Their PnL follows a normal distribution with a mean of 1060.42 and standard deviation of 1835.58. 
Compared to the rational investor, the prospect biased investor (with $c = 2.5$ and $\delta = 0.65$) has lower average daily PnL (117.3) and larger variance ($std=3740.26$). The PnL distribution of the prospect biased investor is also heavily left skewed: there is no gain larger than 3000 and there is a substantial amount of losses around -10000. This is because the prospect biased humans are risk-averse in gains and risk-seeking in losses, which leads to small variance in the positive domain and large variance in the negative domain.
Since $\beta$ in the bounded rationality model \cref{eq:boltzmann} decides how well the agent optimizes their objective function, the PnL increases as $\beta$ increases. We observe that bounded rational agent with zero intelligence ($\beta=0$) has the overall worst average daily PnL = 29.75 and high variance. 
When $\beta$ increases to 1, the human becomes more rational: the average PnL approaches that of the rational investor and the variance reduces.
The myopic investor ($\gamma=0$), on the other hand, has smaller variance (867.96) compared to the fully rational investor, but worse average daily PnL (162.76). The small variance of myopic investor PnL comes from the fact that they only make short-term investments through the trading day, which result in gains and losses in a smaller magnitude compared to long-term investments.
% \kshama{The std of myopic investor PnL maybe small because it's looking only at one-step rewards?}
The PnL of the myopic investor becomes similar to the fully rational investor when $\gamma=0.8$.
The pessimistic investor ($\omega=-1$) has the smallest variance in the PnL distribution (561.98) but low average daily PnL (174.11). This is because they are extremely cautious in investing and often hold a small inventory. Therefore, they are unlikely to have large gains or losses by the end of the day.
However, the optimistic investor has the largest variance in the PnL distribution due to their over-confident and aggressive trading strategy: they can win a large reward if the market goes according to their expectation, or lose a massive amount of money if they made a wrong expectation.

\noindent\textbf{Summary.} We observe that different sub-rational trading strategies have unique daily-PnL distributions due to the noise and bias in the decision-making process. Our experiments also demonstrate that market simulation is an efficient tool to test and evaluate various trading strategies in a controlled environment.

\begin{figure*}[t]
    \centering
    \includegraphics[width=0.44\linewidth]{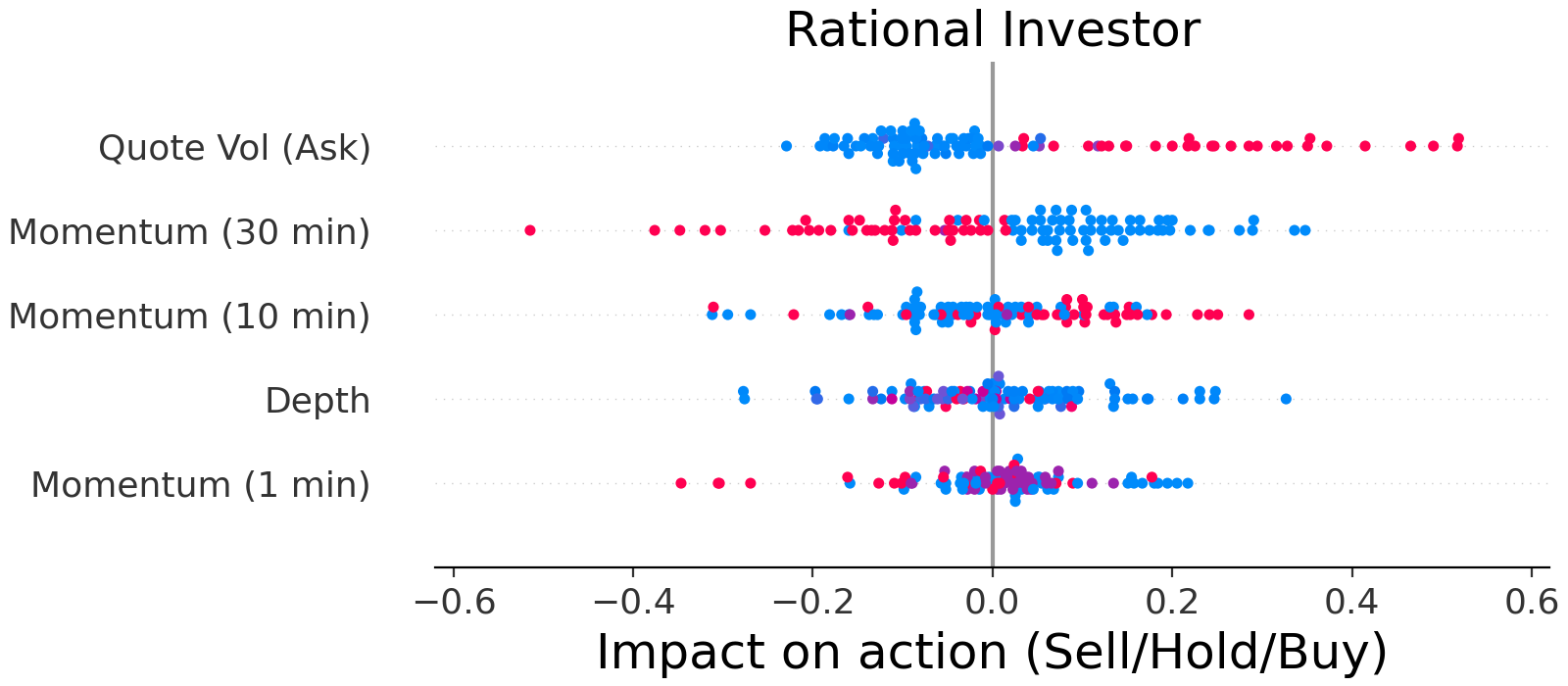}
    \includegraphics[width=0.44\linewidth]{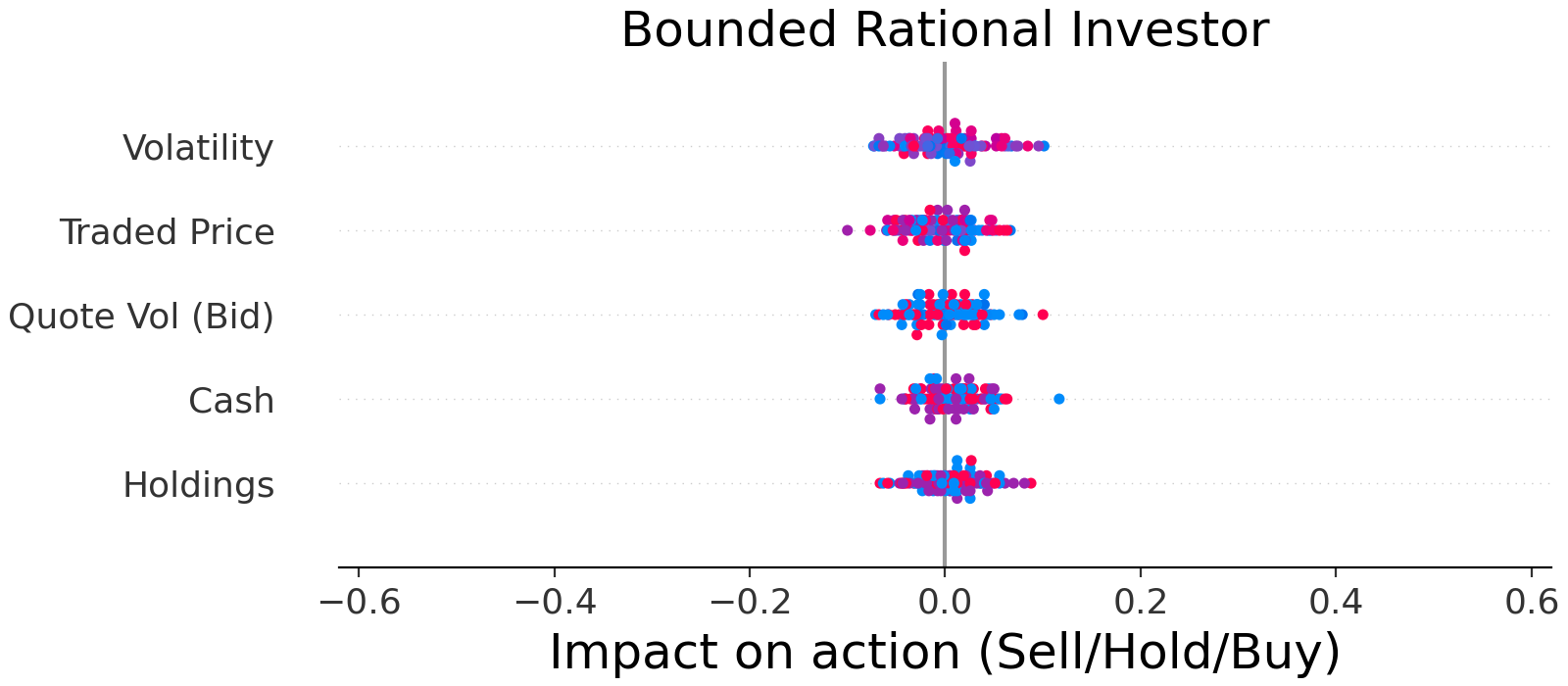}
    
    \vspace{3ex}
    \includegraphics[width=0.44\linewidth]{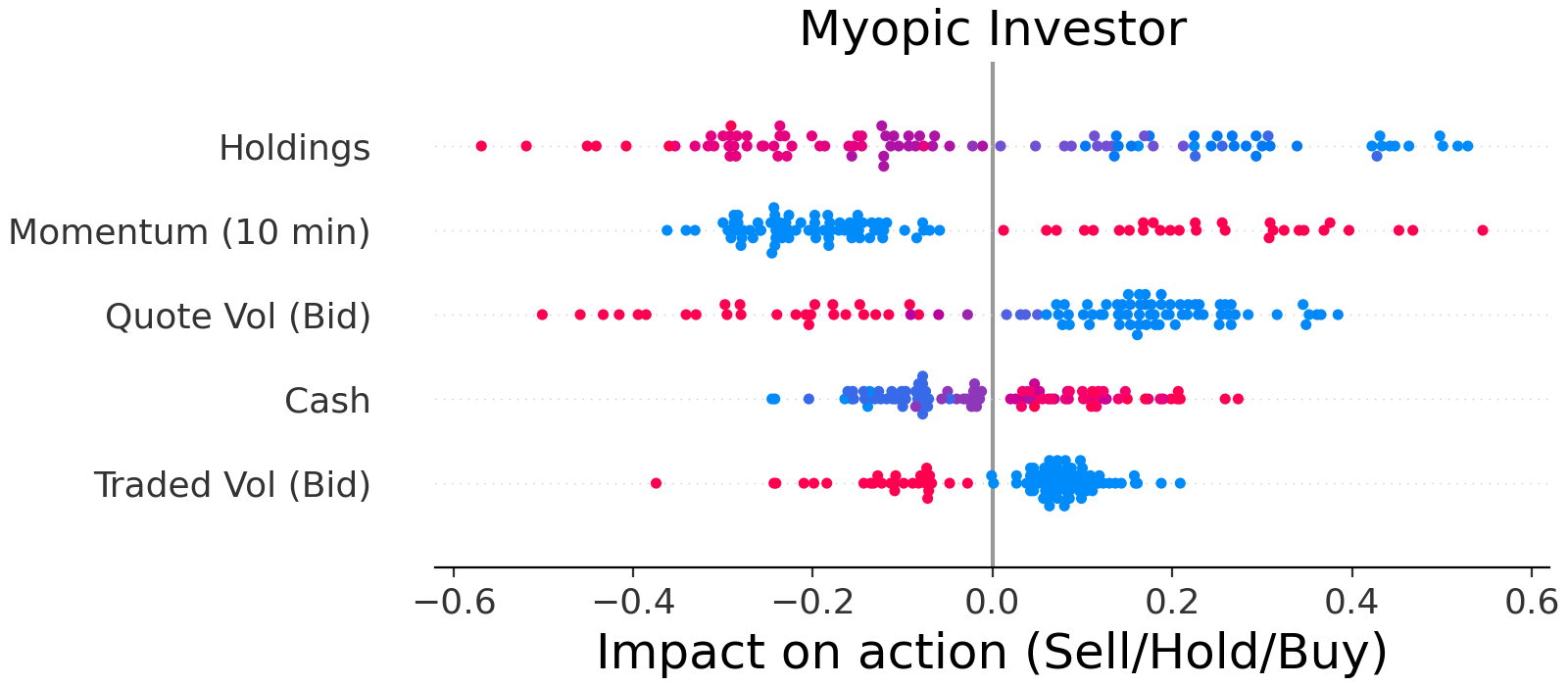}
    \includegraphics[width=0.44\linewidth]{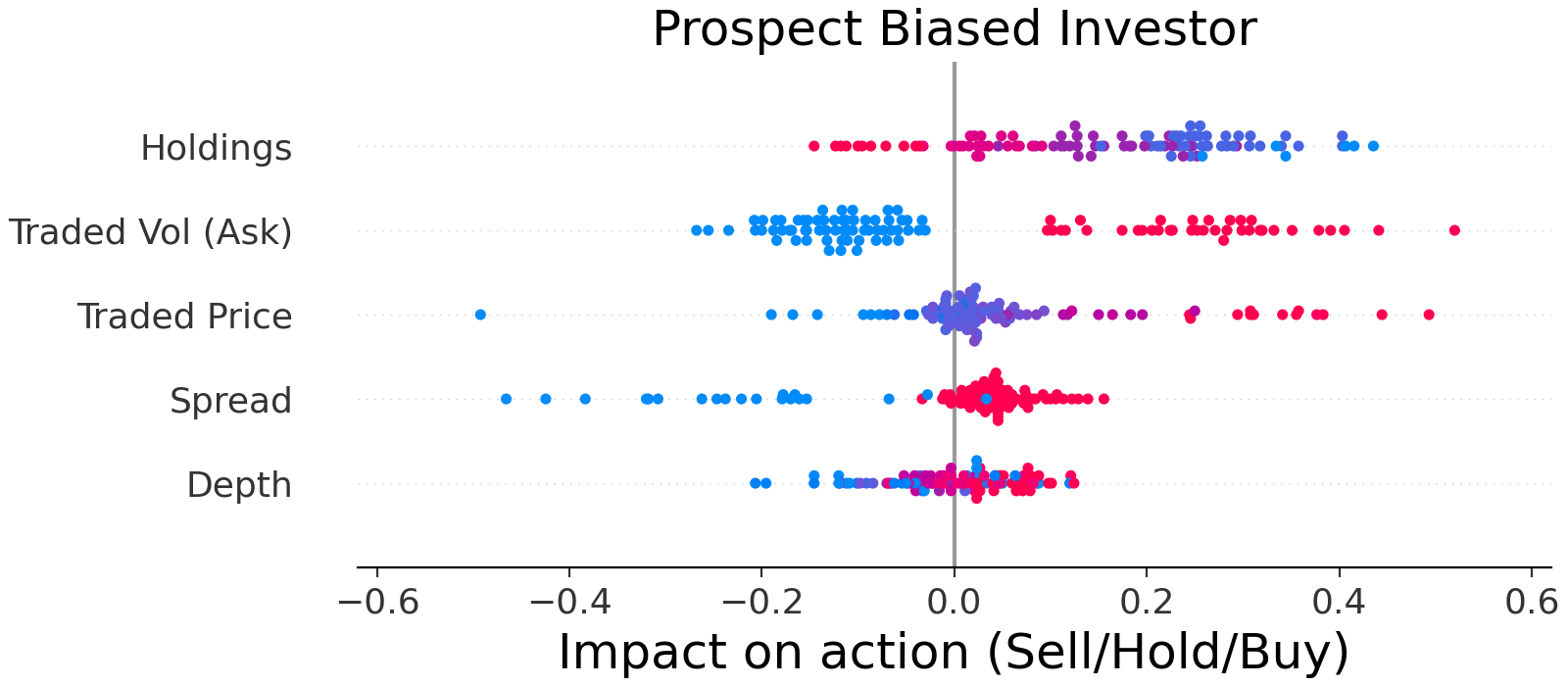}
    
    \caption{The impact of state features on the actions of investor (Sell/Hold/Buy) shown for the top 5 features. 
    The color of the scatter point represents the feature value: blue $\to$ low and red $\to$ high.
    }
    \label{fig:shap}
    \vspace{-1ex}
\end{figure*}

\begin{figure*}[t]
    \centering
    \includegraphics[width=0.44\linewidth]{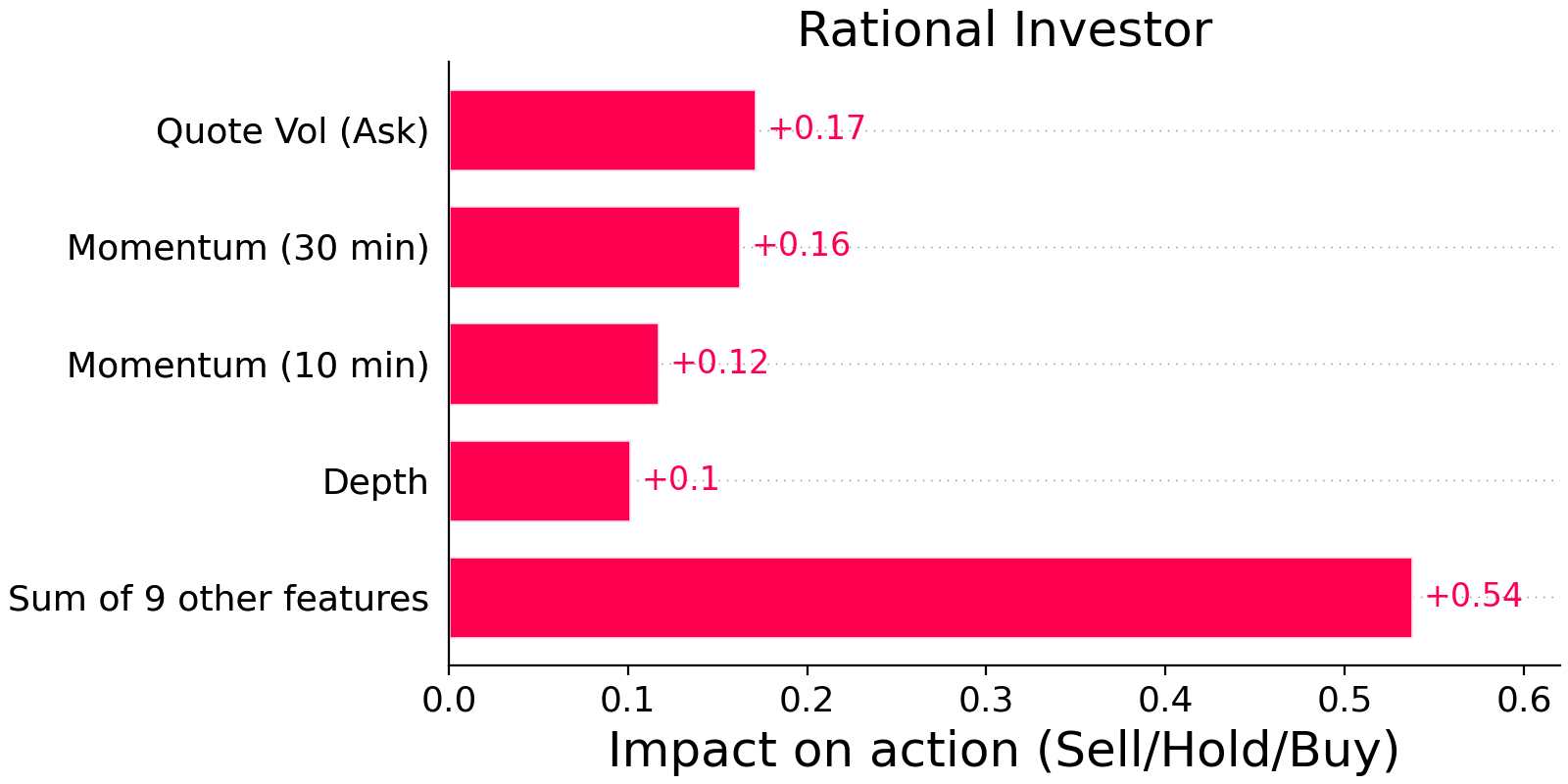}
    \includegraphics[width=0.44\linewidth]{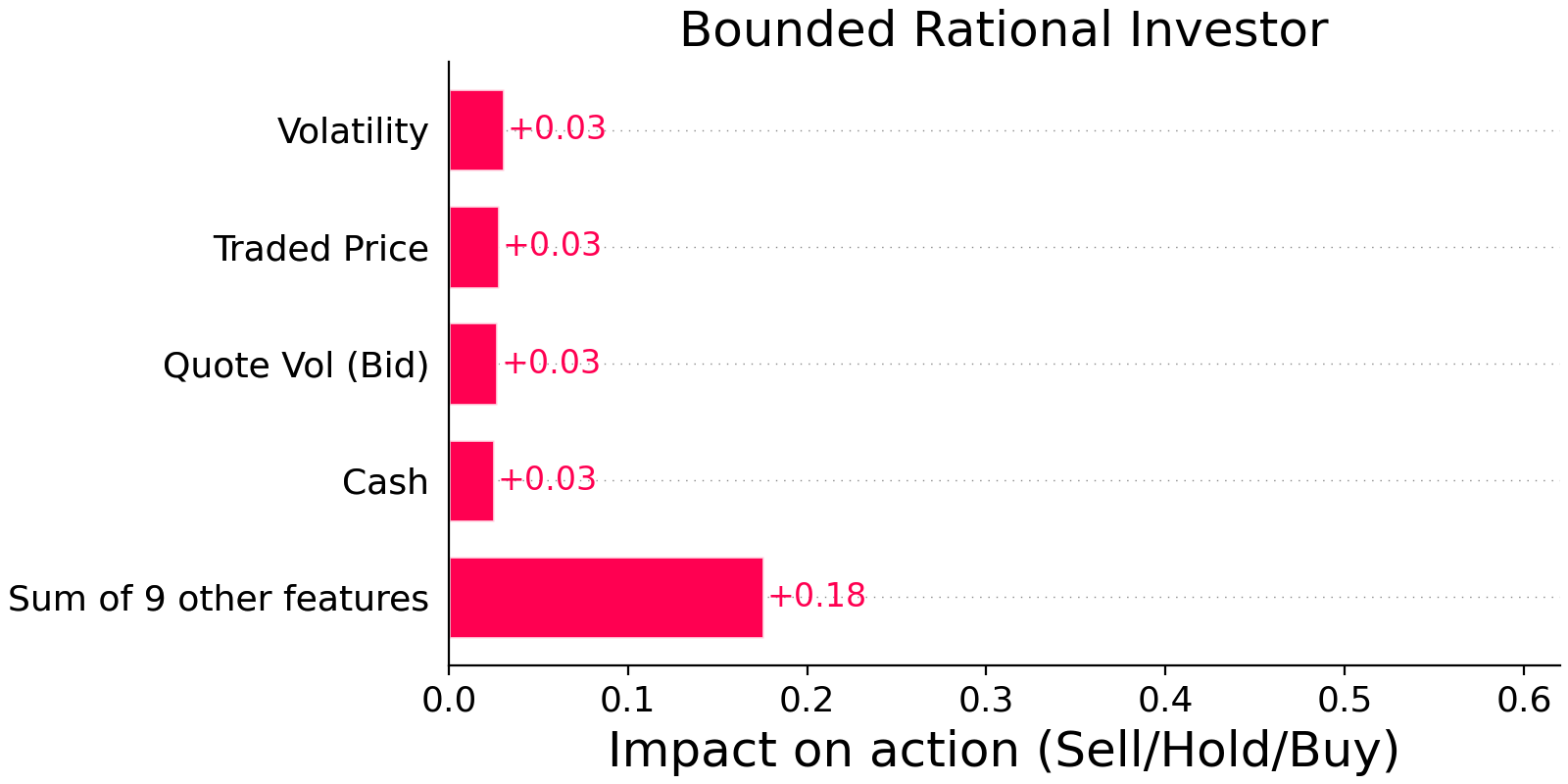}
    
    \vspace{3ex}
    \includegraphics[width=0.44\linewidth]{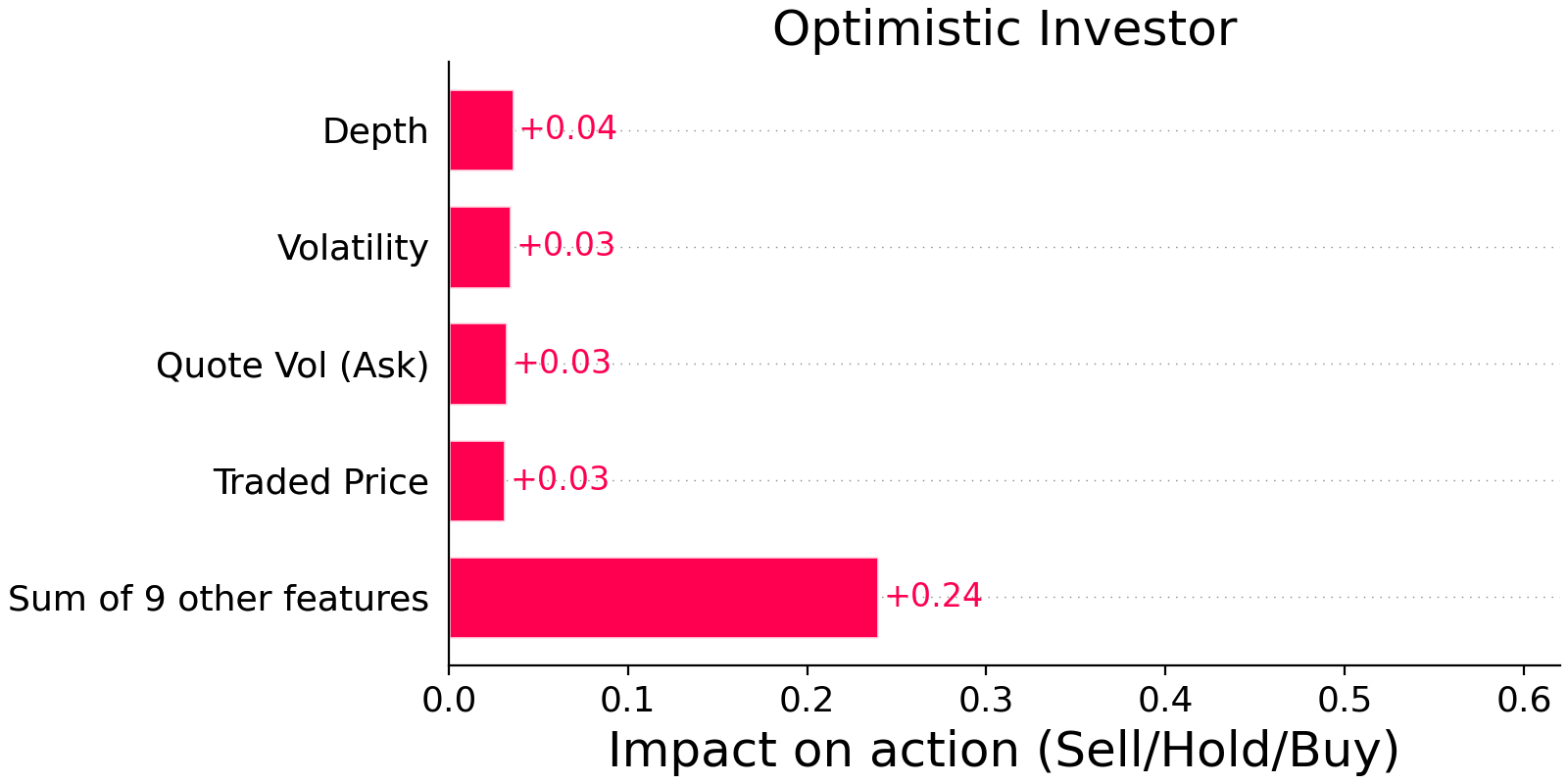}
    \includegraphics[width=0.44\linewidth]{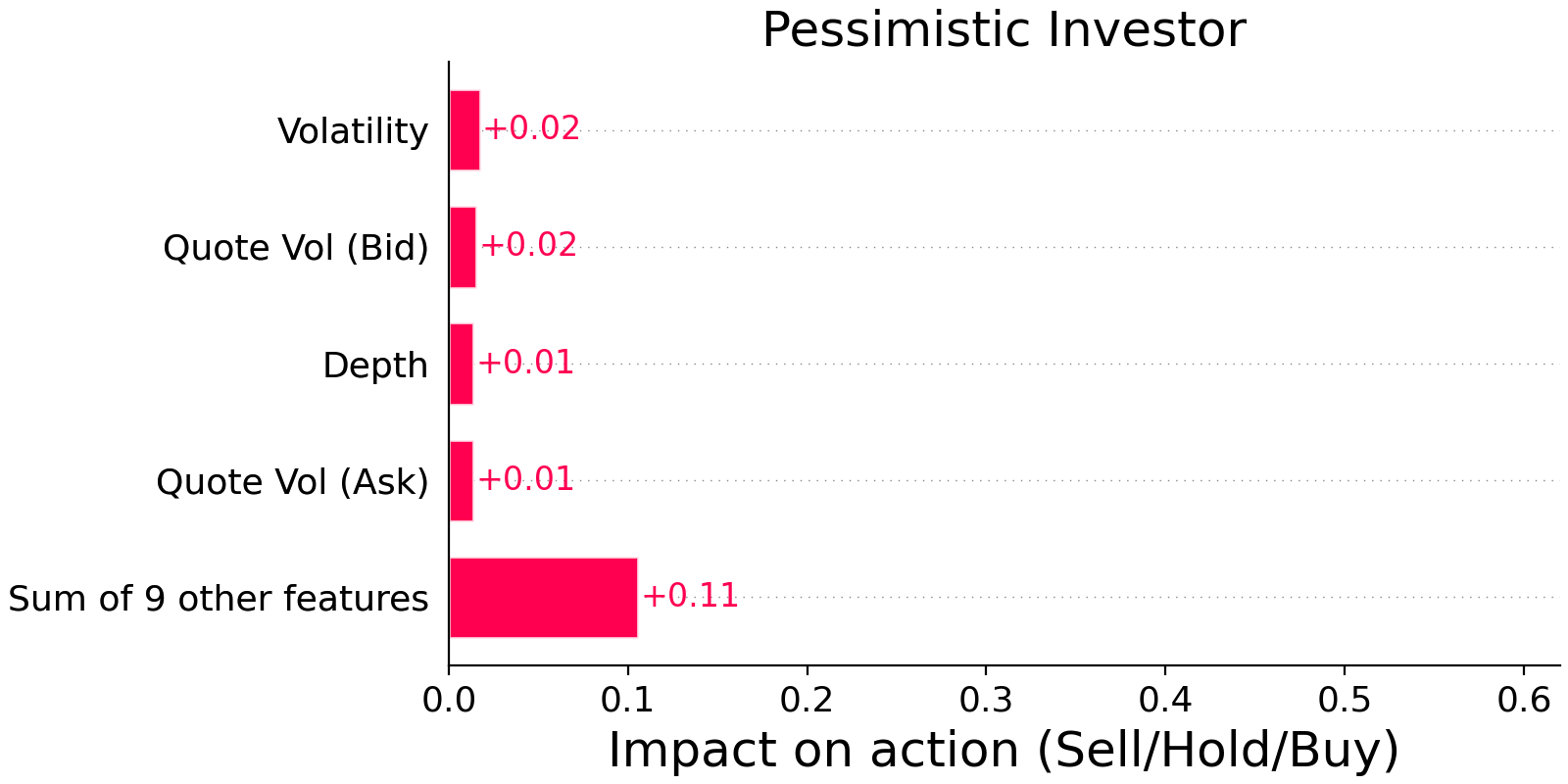}
    
    \caption{Importance of state features towards actions of the sub-rational investor as measured by the average of absolute SHAP values. State features are ordered in decreasing order of their importance on the y-axis. Human investors with bounded rationality or extreme internal beliefs (optimistic/pessimistic) pay less attention to market observations when making decisions.}
    \label{fig:shap_bar}
    \vspace{-1ex}
\end{figure*}
\subsection{Investigation of Sub-rational Human Decision-making}\label{sec:shap}

% \penghang{To Kshama: please feel free to supply any addition information of the SHAP analysis.}
To explain the decision-making process of sub-rational human investors, we perform an explainability analysis of the RL policy network that maps the current state to actions of the investor. We use the explainability tool called SHapley Additive exPlanations (or SHAP in short) to score the different state features based on their importance towards determining the trader's actions (\cite{lundberg2017unified}). These scores, called SHAP values, are computed using cooperative game theory by decomposing the
network output locally into a sum of effects attributed to each input feature. Hence, every (state, action) pair has a vector of SHAP values with elements corresponding to each state feature, and that accumulate to give the action per action dimension.
% Here we use the SHapley Additive exPlanations (SHAP) (\cite{lundberg2017unified}) to examine the sub-rational policies. 
% Based on Shapley values from cooperative game theory, the SHAP method is designed to provide a consistent and fair way of attributing the contribution of each feature to the prediction made by a model. Applying SHAP analysis to state-action relations in RL can provide insights into the importance of different states in determining the choice of actions by the RL agents.

We compute SHAP values for every entry in a dataset comprising (state, action) pairs for the different sub-rational investor policies.
\cref{fig:shap} shows the distribution of SHAP values for each state feature towards determining the order direction, ordered by the corresponding feature importance (sum of absolute values of SHAP values). The x-axis indicates the impact on the model output (-1$\to$sell, 0$\to$hold, 1$\to$buy) while the color represents the observed value of the state (blue$\to$ low, red$\to$ high). 
For example, the 30-minute momentum is one of the most impactful state features for the fully rational policy. If the momentum is low (blue), it has a positive impact on the investor's action and they place a buy order, and vice-versa when the momentum is high (red). So, the rational agent will place buy/sell orders if they observe a strong decrease/increase in the past 30 minutes, indicating that the current price is at a low/high position. 
% For example, the 30-minute momentum is the most impactful state in the fully rational policy: the rational agents will place sell/buy orders if they observed a strong increases/decreases in the past 30 minutes, indicating the current price is at a high/low position. 
The short-term momentum is less important to the rational investor, especially for the 1-minute momentum which shows a hardly distinguishable SHAP value distribution.

For the zero-intelligence bounded rational investor ($\beta=0$), we do not observe a distinguishable contribution of the state features to the agent's actions. The distribution of SHAP values is not correlated with the value of the state features.
Compared to the rational investor, the myopic investor ($\gamma=0$) focuses more on the short-term market observations, as the feature importance of 10-minute momentum is higher than the 30-minute momentum. They are more careful on their current portfolio situation (holding, cash) opting to sell when holdings are high, and buy when holdings are low, which explains the myopic loss aversion behavior.
The SHAP-value analysis of the prospect biased investors shows that holding is the most important state feature for their decision-making. This indicates that the prospect biased investors are very sensitive to the inventory risk (\cite{avellaneda2008high}). When experiencing gains in the market, they incline to reduce the inventory if they have a large amount of holdings.

We also notice that investors with bounded rationality ($\beta=0$), pessimistic bias ($\omega=-1$), or optimistic bias ($\omega=1$) do not build their decisions based on any market observations. As shown in \cref{fig:shap_bar}, none of the state features gives average SHAP values greater than 0.05 for bounded rational, pessimistic, and optimistic investors, which suggests that they are not strong contributing factors for the investors' decisions.

\noindent\textbf{Summary.} The SHAP value analysis of the state-action relationship gives an explicit explanation of the sub-rational human strategy. We observe that the zero-intelligence bounded rational investors as well as those with extreme internal bias (optimism/pessimism) do not make decisions based on their observations of the markets, the myopic investors pay more attention to the short-term momentum, and the decisions of prospect biased investors strongly base decisions on their inventory --  which is aligned with one's expectations about these types of sub-rational behaviors.

\subsection{Impact of Sub-Rational Investors on Market Observables}
\label{sec:impact_of_human}

\begin{figure*}[t]
    \centering
    \includegraphics[width=0.24\linewidth]{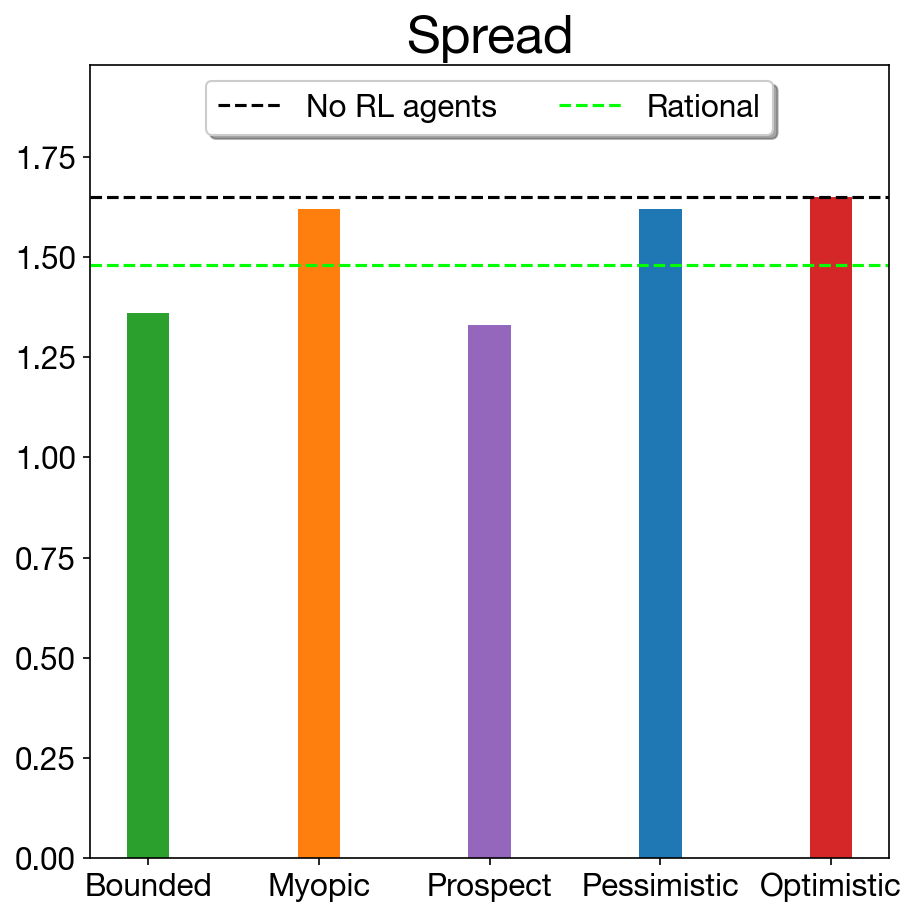}
    \includegraphics[width=0.24\linewidth]{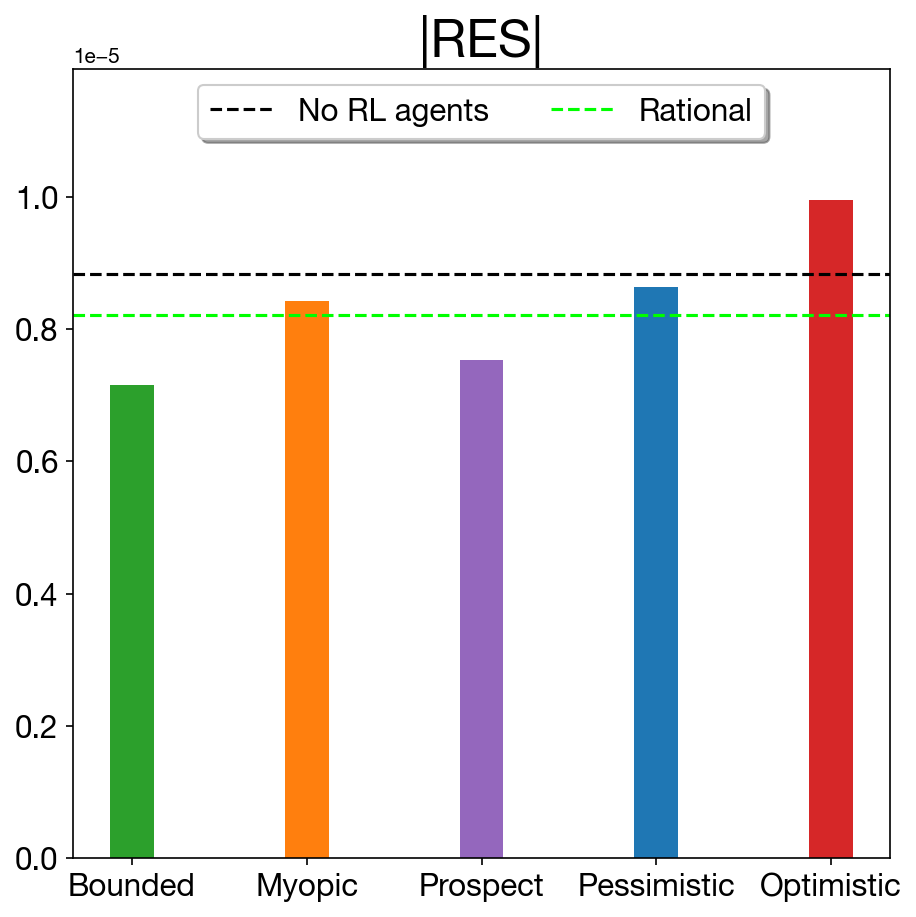}
    \includegraphics[width=0.24\linewidth]{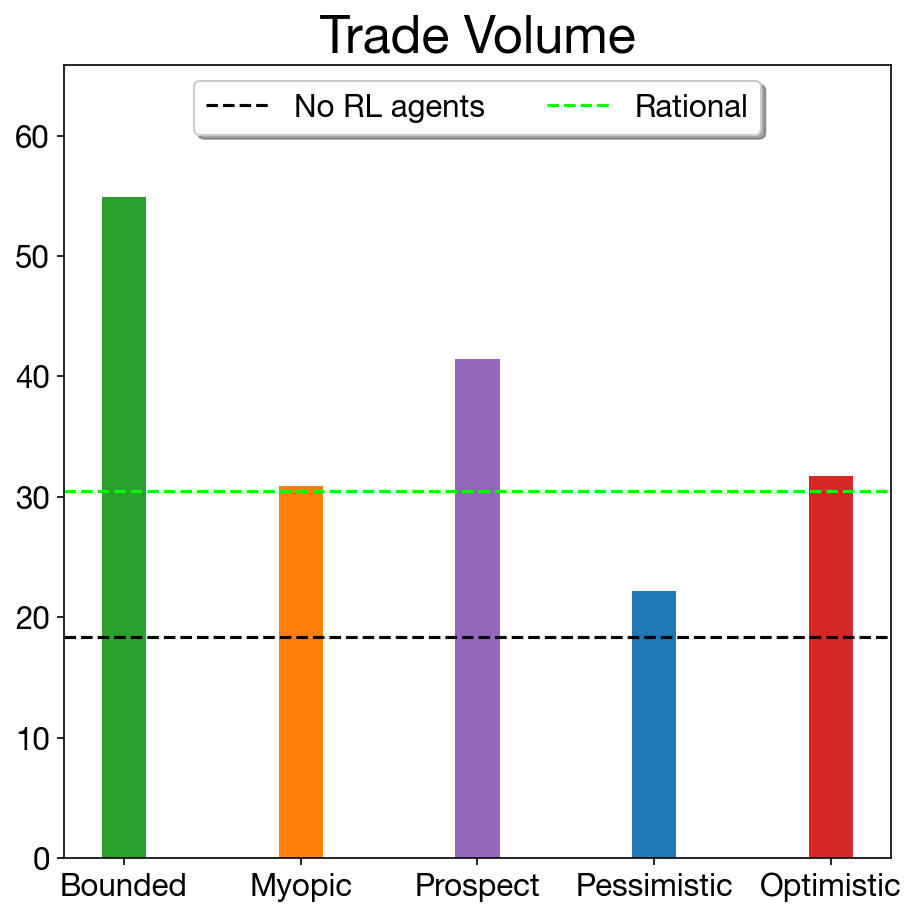}
    \includegraphics[width=0.24\linewidth]{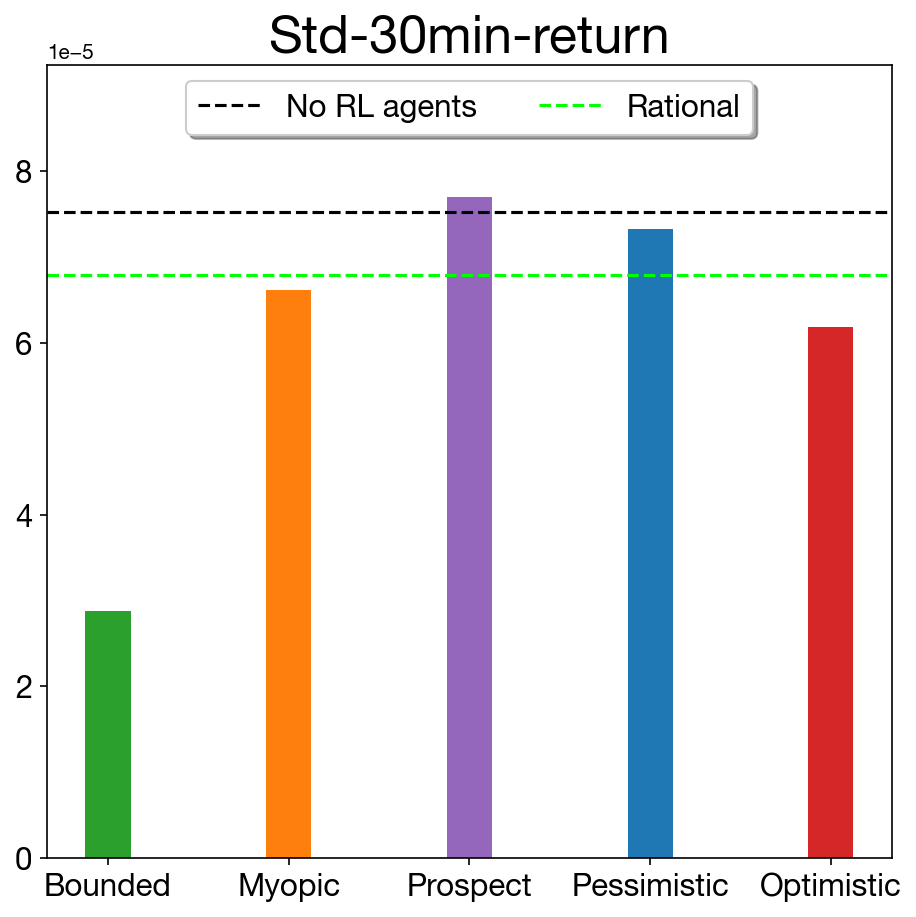}
    
    \vspace{3ex}
    \includegraphics[width=0.24\linewidth]{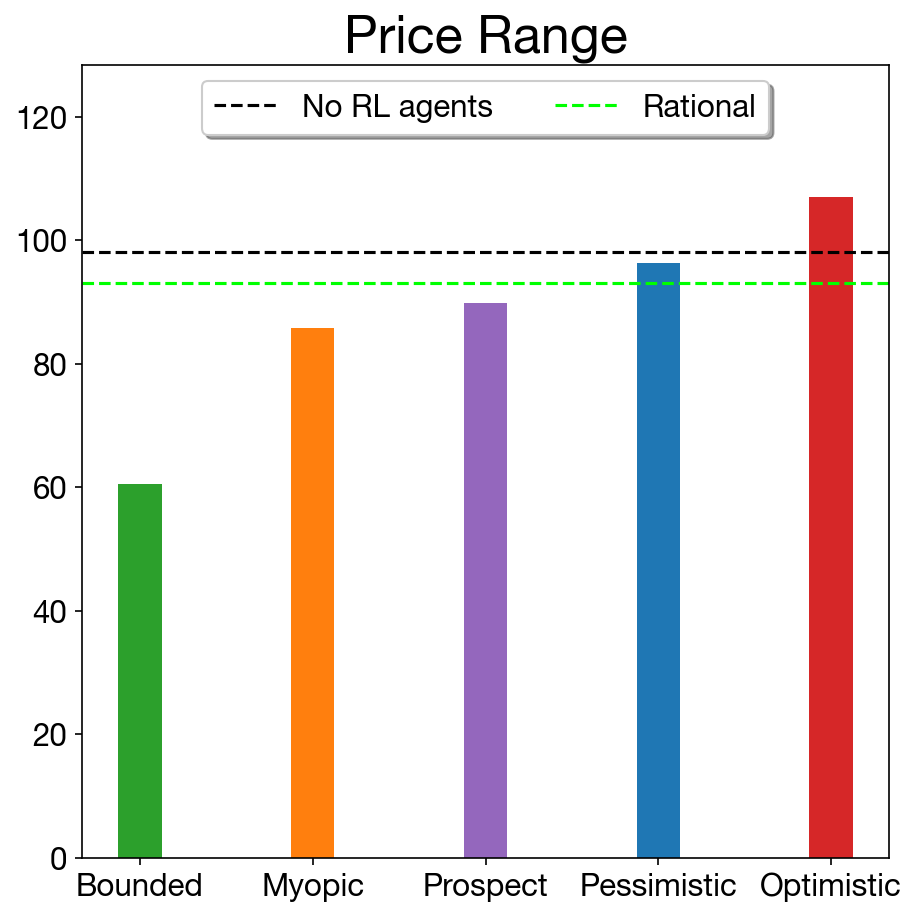}
    \includegraphics[width=0.24\linewidth]{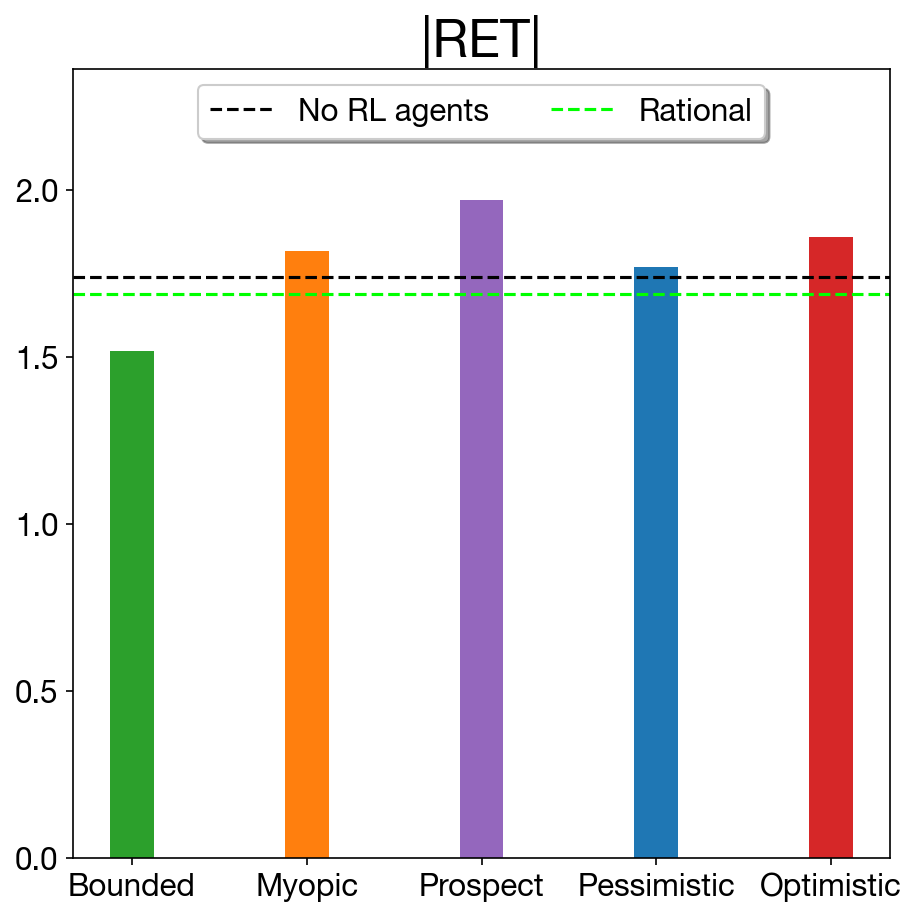}
    \includegraphics[width=0.24\linewidth]{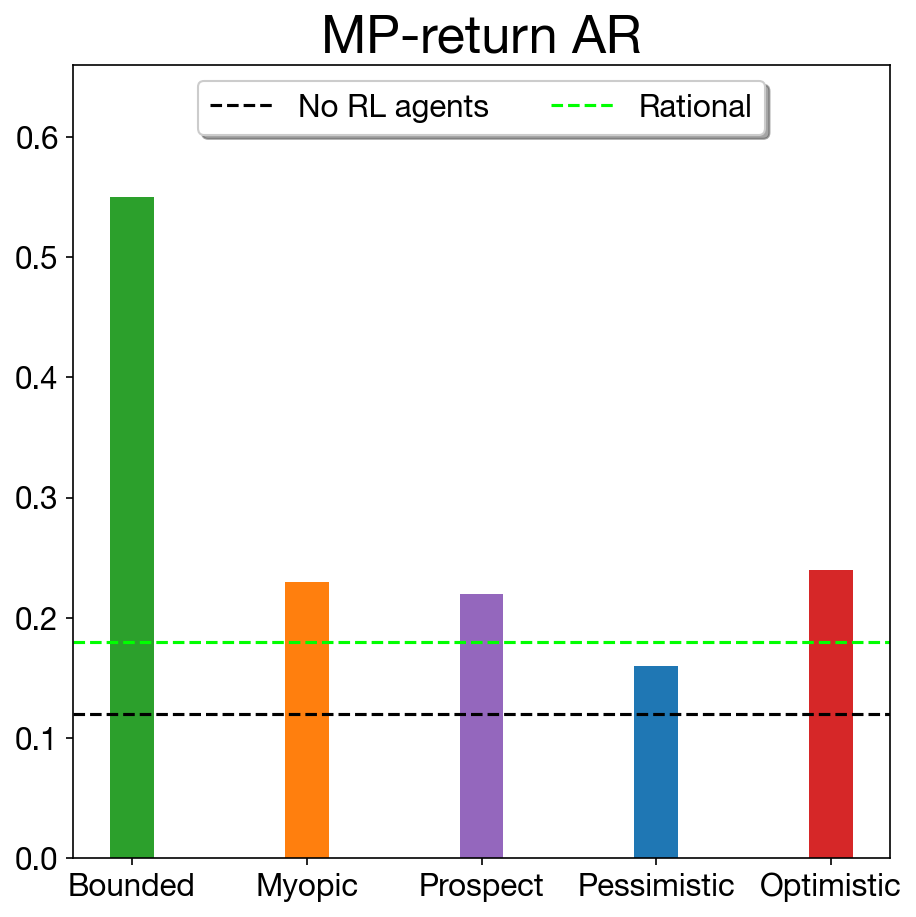}
    \includegraphics[width=0.24\linewidth]{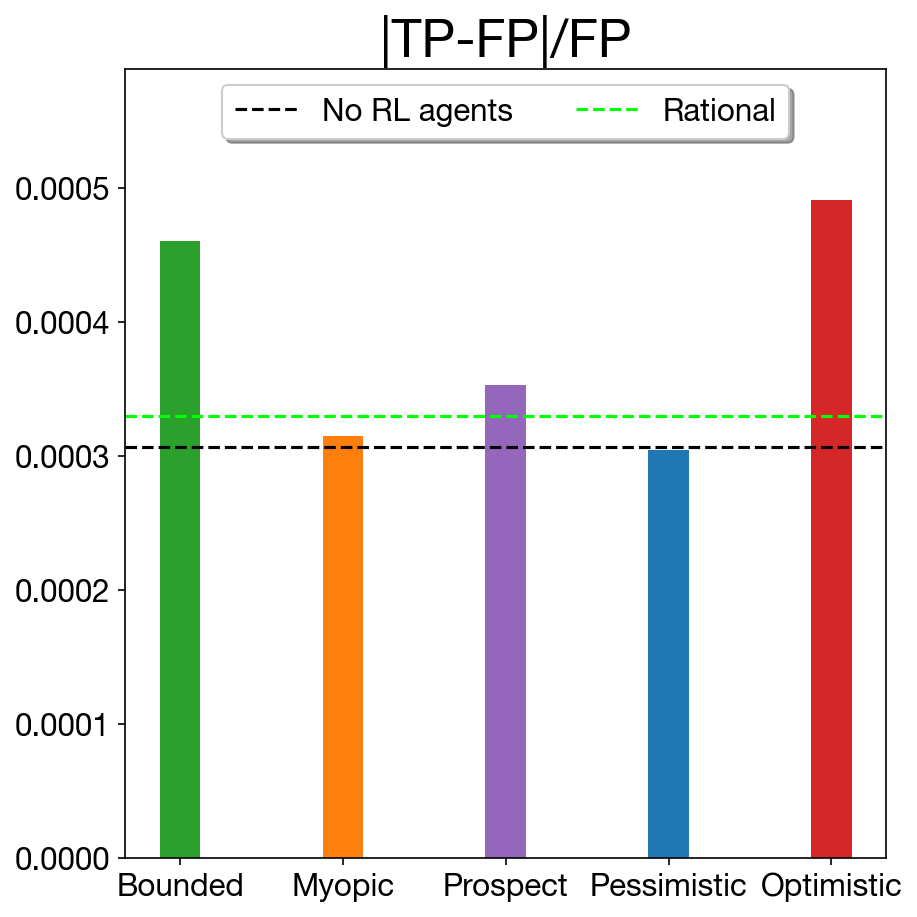}
    
    \caption{Impact of different types of RL investors on market quality. As mentioned in \cref{sec:quality_metrics}, we evaluate market quality on three aspects: liquidity (spread, $|$RES$|$, trader volume), volatility (std-30min-return, price range, $|$RET$|$), and efficiency (MP-return AR, $|$TP-FP$|$). }
    \label{fig:market_impacts}
    \vspace{-1ex}
\end{figure*}

Here, we investigate how sub-rational trading behaviors affect the financial market. In particular, we examine and compare the statistics of simulated markets with different types of human investors while the background agents configuration (see \cref{tab:configuration}) remains the same. 
We perform this exercise for seven different market settings: no RL agents, rational (10 rational RL agents), bounded-rational (10 RL agents with $\beta=0$), psychologically myopic (10 RL agents with $\gamma=0.2$), prospect biased (10 RL agents with $c=2.5$ and $\delta=0.65$), optimistic (10 RL agents with $\omega=1$), and pessimistic (10 RL agents with $\omega=-1$).
\cref{fig:market_impacts} shows average metrics per trading day over 100 simulations for each market setting described earlier in this section. The simulation random seeds are different per simulation run within each setting and same across different settings. 
We evaluate market quality on dimensions of liquidity, volatility, and information efficiency as described in \cref{sec:quality_metrics}. For all metrics except trade volume, smaller values indicate better market quality.
We observe that market liquidity almost always improves with the addition of RL investors except in the optimistic setting for $|$RES$|$.

\begin{figure*}[t]
    \centering
    \includegraphics[width=0.69\linewidth]{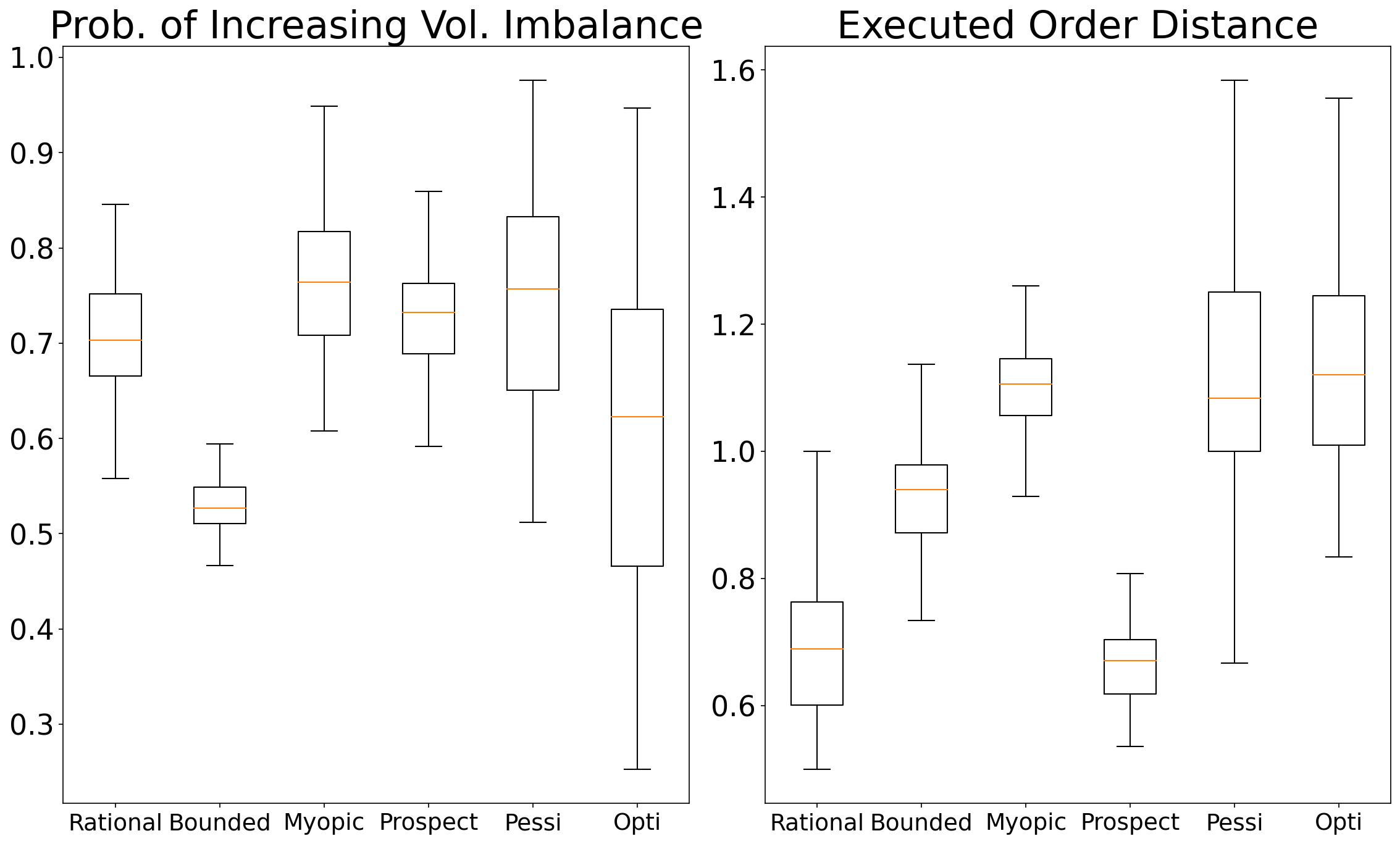}
    
    \caption{The distribution of agents' actions in simulated markets. (left) The probability of the agent placing orders on the heavier side of the LOB, which increases the volume imbalance. (right) The distance between the price of executed orders and the mid price.
    }
    \label{fig:action_details}
    \vspace{-1ex}
\end{figure*}

\noindent\textbf{Bounded rational investor.} We observe that the bounded rational investor leads to markets with better liquidity and volatility, but worse information efficiency. Compared to simulated markets with rational investors (rational) and without any RL investor, markets with bounded rational investors have larger trade volume and smaller spread, $|$RES$|$, Std-30min-RET, price range, and $|$RET$|$. 
This arises from the fact that a collection of zero-intelligence bounded rational investors essentially mirrors the functions of a market maker \cite{venkataraman2007value}. While a collection of bounded rational investors tend to place orders uniformly at random on both sides of the LOB due their limited computational capacity, market makers are required by regulation to place orders on both sides regularly. 
% This arises from the fact that a collection of zero-intelligence bounded rational investors essentially mirrors the functions of a market maker: they tend to place orders randomly on both sides of the LOB. 
Additionally, as illustrated in \cref{fig:action_details}, bounded rational investors exhibit a lower likelihood of placing orders on the side of the LOB with larger quoted volume, and the orders they execute are in proximity to the mid-price. Accordingly, due to their limited utilization of market information as seen in \cref{fig:shap}, the bounded rational investors noticeably diminish market efficiency.

\noindent\textbf{Myopic investor.} As we discovered in the SHAP analysis (\cref{fig:shap}), the myopic investor's decisions tend to follow the short-term market momentum. As a result, the myopic investor typically positions orders on the side of the LOB with larger quoted volume, thereby increasing the LOB imbalance. Our observations in \cref{fig:market_impacts} indicate that myopic trading behavior has negative impacts on market liquidity when contrasted with rational behavior. However, it enhances market efficiency with regard to the $|$TP-FP$|$ / FP since it elevates the propagation of fundamental information by following the market momentum. Meanwhile, there is no significant discernible difference in the impact of myopic investors on volatility compared to their rational counterparts.

\noindent\textbf{Prospect biased investor.} The prospect biased investors significantly improve market liquidity by narrowing the spread and $|$RES$|$, and increasing the trade volume. A major reason being that among all types of RL investors, the prospect biased investor has executed orders closest to the mid price as seen in \cref{fig:action_details}. 
This is a result of them placing orders close to the mid-price so as to reduce any uncertainty in order execution. This subsequently reduces the spread and facilitates trade executions.
% This means that prospect biased investors place orders close to the mid prices thereby reducing the spread, and facilitating trade executions.
% This is likely to reduce the gap between sells and buys, and facilitate trade executions.

\noindent\textbf{Pessimistic investor.} Since pessimistic investors are not willing to trade most of the time, simulated markets with pessimistic investors have less liquidity than those with other types RL investors (excepting the optimistic). On the other hand, they do not exhibit strong influences on the market, resulting in the market statistics being very close to those of markets without any RL investors.

\noindent\textbf{Optimistic investor.} As shown in \cref{fig:shap_bar}, optimistic investors rely more on their internal beliefs than market variables, which undermines market efficiency. The aggressive trading strategy of optimistic investors also leads to simulated markets with largest spread, price range and absolute daily returns $|$RET$|$. As illustrated in \cref{fig:opti_pessi}, the actions of optimistic investors are often heavily skewed to buy (or sell). Therefore, they tend to increase the LOB imbalance and reduce the liquidity of the market. 
% \kshama{Not sure if order direction/imbalance is captured by market liquidity metrics.} 

\noindent\textbf{Summary.} By comparing with market simulations without any RL investors, we show that bounded rational and prospect biased investors improve liquidity, but reduce efficiency; myopic investors reduce liquidity but improve efficiency; optimistic investors increase volatility and reduce efficiency; and pessimistic investors do not have strong impacts on the market. All of our aforementioned findings on the different sub-rationality models are summarized in \cref{tab:summary}.

\section{Discussion}

One of the largest obstacles in studying sub-rational trading behavior is the scarcity of historical trading data at an individual level, which is usually not public-available (\cite{gutierrez2019mapping,cont2023analysis}). 
While several studies have collected sub-rational behavior from controlled lab experiments on human subjects (\cite{mischel1970attention,rothblum1986affective,green1997rate}), such data are very limited at scale and may raise ethical and security concerns in the financial domain.
Even with access to labeled trading data, factorizing the impact of human traders on complex market environments remains a challenge.
Recent studies have employed large language models (LLM) to generate synthetic human demonstration data~\cite{coletta2024llm}, which have not been successfully applied for generating complex investing decisions in financial markets. 
As a result, there lacks studies which can identify sub-rational trading behaviors and evaluate their impacts on financial markets.
Among the few that exist, \cite{olsen1998behavioral} gives behavioral finance explanations of market volatility, and \cite{levy2000microscopic} discover that human traders influenced by prospect bias generate price deviation, high traded volume, and excess volatility. Our experimental results in \cref{fig:market_impacts} align with their findings. 
By utilizing RL and market simulations, our model can generate sub-rational trading decisions with or without using ground truth historical data, which liberates research from the scarcity of sub-rational trading data.

While we provide a mathematical framework to capture various aspects of human sub-rationality, it is crucial to specify the sub-rationality parameters for real humans (e.g. $\beta$ for bounded rational humans, or $\gamma$ for myopic humans). A common approach is to estimate these parameters using human demonstration data. For example, \cite{tversky1992advances} used a nonlinear regression model to estimate the parameters in \cref{eq:prospect_utility,eq:prospect_prob} from human subject studies. Similarly, \cite{green1997rate} collected decisions of human subjects and fit exponential and hyperbolic discounting factors (\cref{eq:exponential,eq:hyperbolic}) using nonlinear regression. 
However, these sub-rationality parameters vary between humans with different degrees of bias and intelligence, making it expensive to conduct human studies more broadly.
Consequently, existing works on estimating parameters for human models show disagreement among each other (\cite{frederick2002time}).
Hence, we do not aim to estimate a single subrationality parameter for humans using trading data, and instead propose a holistic framework to model human investors with different types and degrees of sub-rationality. And, investigate the impact of such subrationality on trading behaviors and market environments. 
\section{Conclusion}
In this work, we employ reinforcement learning (RL) techniques to model sub-rational human investors in financial markets. We consider sub-rational behavior as explained by two driving factors: limited information/computational capacity and biased internal beliefs. Our approach involves utilizing sophisticated market simulations for the training and evaluation of diverse human investor models. Additionally, we construct probabilistic neural networks to accurately represent the biased internal beliefs guiding human investors in their decision-making process.

In particular, we consider five sub-rational human investor models:
(1) the bounded rational investors that make sub-optimal decisions due to limitations in information access and computational power, (2) the psychologically myopic investors that maximize only short-term rewards regardless of the future, (3) the prospect biased investors that are risk-averse in gains and risk-seeking in losses, (4) the optimistic investors that have exaggerated hope in receiving positive outcomes, (5) and the pessimistic investors that tend to underestimate the likelihood of receiving positive outcomes over negative outcomes.
We first craft specific market scenarios to assess and illustrate the behavior of each type of sub-rational human investors.
We then examine the relationship between the Profit and Loss (PnL) of these investors and their respective computational limitations and psychological biases. Furthermore, we provide an intuitive exploration of the sub-rational human trading strategies through SHAP value analysis. We show that our models successfully reproduce sub-rational trading behavior identified in the previous literature.
Upon populating simulated markets with different types of sub-rational human investors, we evaluate their impact on market quality in terms of liquidity, volatility, and price efficiency.
Our findings indicate that a market populated with bounded rational or prospect biased investors has improved liquidity but worse price efficiency. Additionally, we observe that bounded rational investors contribute to a reduction in stock volatility. On the other hand, myopic investors show adverse impacts on liquidity but improve information efficiency. For the pessimistic trading behavior, we do not identify a significant impact on the market quality. Meanwhile, the optimistic investors cause negative impact on market liquidity and price efficiency.
% We found that the bounded rational investors can increase the traded volume and reduce the spread and volatility. 
We note that our comparative study is different from previous regulatory work on algorithmic and high-frequency trading \cite{algo_trading, woodward2017need} in that we compare investors with different types of rationality, while not varying their trading frequencies in simulated markets.
% Focusing on the rationality of investors, our models provide new insights to the impact of human investors on markets apart from studies in algorithimc trading and high frequency trading\cite{algo_trading, woodward2017need}. 
%Although we expected that sub-rational human investors can change the market environment, we did not observe significant influence of the bounded rational investors.

A future direction of this work would be to consider other aspects of human sub-rationality. For example, economic studies show that investors make sub-optimal decisions due to overconfidence, which can be described as resulting from an illusion of knowledge or an illusion of control \cite{barber2002online, song2013overconfidence}.
In addition, we are interested in estimating the degree of sub-rationality for real human investors, using historical data and (potentially) demonstration data generated by large language models as proxy human subjects \cite{coletta2024llm}. 
% \kshama{To add Paul Nixon's work at ICAIF once it's available publicly, related to modeling outcome expectation bias of humans.}
% In addition, we plan to explain the human investor policy with respect to the features in the state space, which we believe will provide better understanding of the sub-rational trading behaviors. 
% \cite{barber2002online} economy study describes overconfidence in investing as illusion of knowledge and illusion of control.
% \cite{song2013overconfidence} indicates that overconfidence in one’s ability leads to suboptimal actions, in particular in financial decision making. They show that negative returns of previous investments lead to lower investment incidence, lower investment in principal-unprotected products, and lower investment quantity. However, overconfident investors are less sensitive to such negative returns and achieve enhanced investment returns compared to less confident investors.
% \section{Acknowledgements}
%%
%% The next two lines define the bibliography style to be used, and
%% the bibliography file.

\section*{Acknowledgements}
This paper was prepared for informational purposes by the CDAO group of JPMorgan Chase \& Co and its affiliates (``J.P. Morgan'') and is not a product of the Research Department of J.P. Morgan. J.P. Morgan makes no representation and warranty whatsoever and disclaims all liability, for the completeness, accuracy or reliability of the information contained herein. This document is not intended as investment research or investment advice, or a recommendation, offer or solicitation for the purchase or sale of any security, financial instrument, financial product or service, or to be used in any way for evaluating the merits of participating in any transaction, and shall not constitute a solicitation under any jurisdiction or to any person, if such solicitation under such jurisdiction or to such person would be unlawful.
\pagebreak
\bibliographystyle{tfcad}
\bibliography{main}

\pagebreak
\section{Appendices}
\subsection{Performance of Internal Belief Models} \label{sec:internal_validation}
\begin{figure*}[h]
    \centering
    \includegraphics[width=0.45\linewidth]{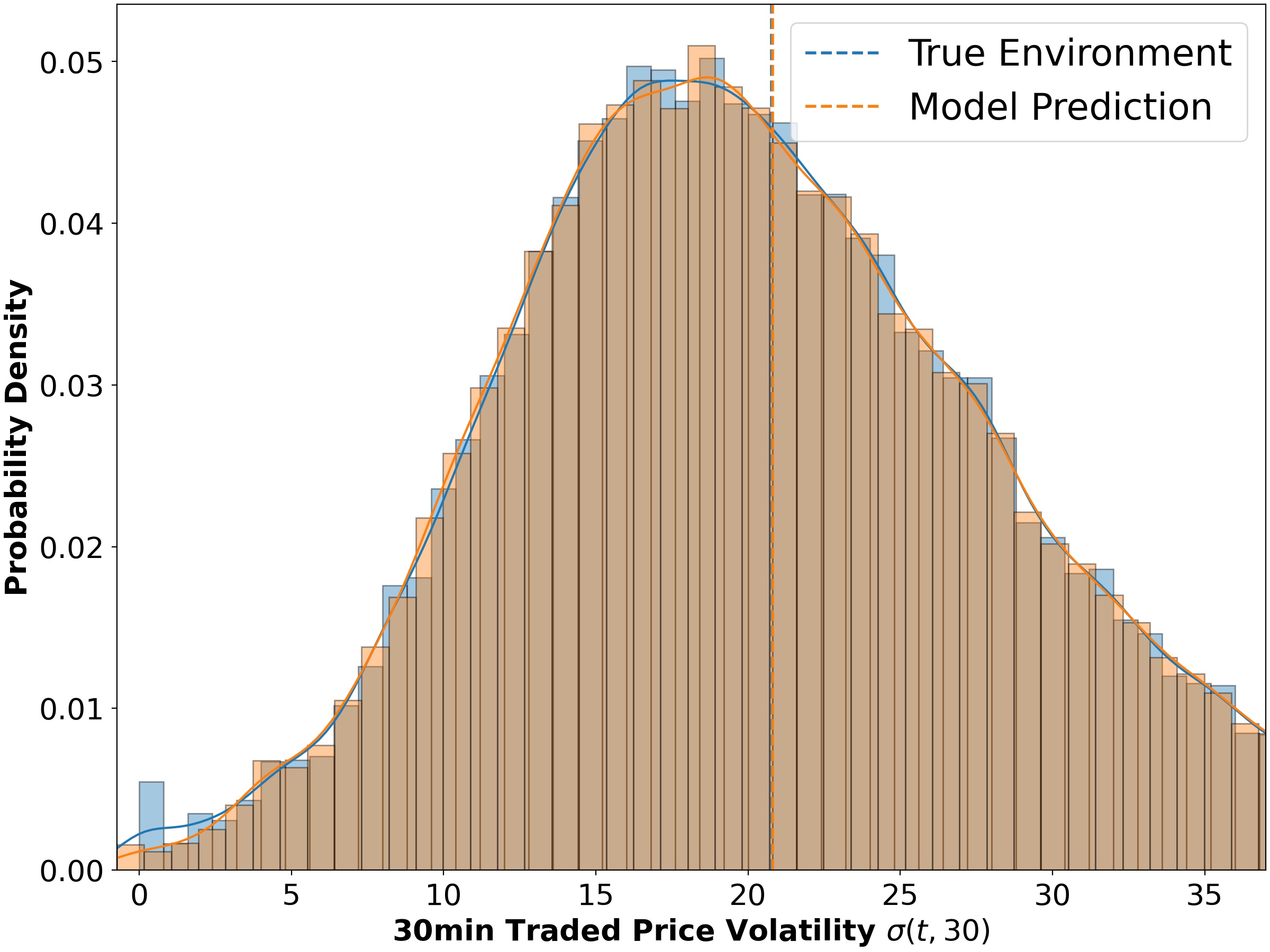}
    \includegraphics[width=0.45\linewidth]{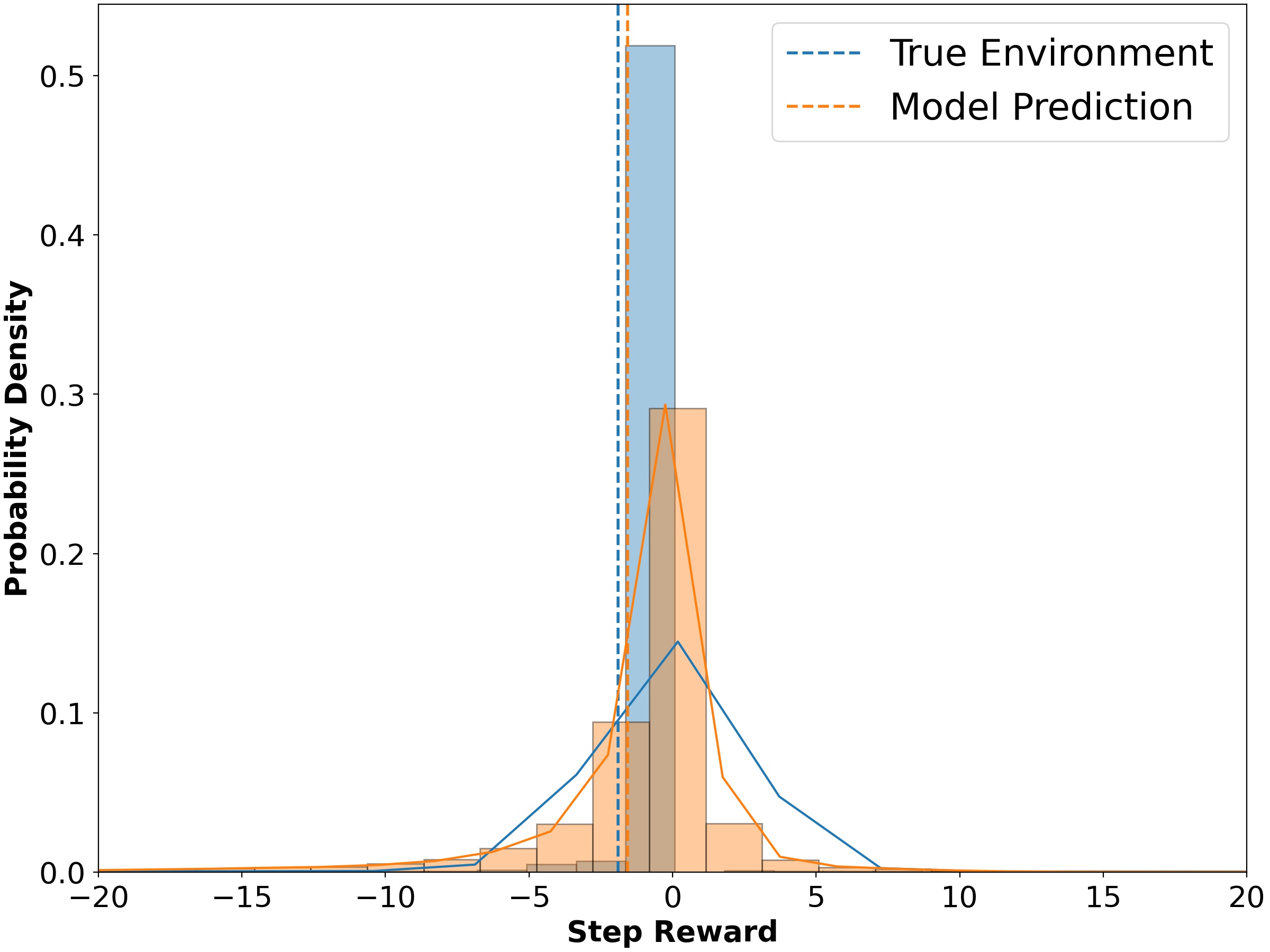}
    \caption{The distributions of 30-minute traded price volatility and step rewards from the internal model and the true environment.}
    \label{fig:model_dist}
    \vspace{-1ex}
\end{figure*}

% \noindent\textbf{Probabilistic Neural Networks}
As illustrated in \cref{sec:internal_model}, we use probabilistic neural networks (PNN) as the internal model to incorporate biased human beliefs. To evaluate the performance of the internal model, we generate sequences of random trading actions and feed them to the true environment (ABIDES) and internal model (PNN) respectively. \cref{fig:model_dist} shows the distributions of 30-minute traded price volatility and the step reward with mean and kernel density estimation curve. In addition, \cref{tab:model_compare_stats} shows the Earth Mover's Distance (EMD, also known as Wasserstein distance) and the root mean square error (RMSE) between the internal model prediction and the true environment outputs. The results show that the internal model can be considered as an alternative of the true environment for injecting the biased human beliefs. Note that the error for spread and volume is slightly higher. This discrepancy may arise from the fact that PNN generates the predictions from Gaussian distributions, which might not be an ideal fit for variables following heavy-tailed distributions. As a part of our future work, we intend to enhance this aspect by developing a more sophisticated world model.

\begin{table}[b]
    \vspace{-3ex}
    \centering
    \caption{The distribution of the internal model vs. the true environment. Each variable is normalized by the min-max values of the combined distributions of PNN and the true environment.}
    \begin{tabular}{lrr}
    \hline
Variable	&	EMD	&	RMSE \\ \hline
Quote Volume$_{t+1}$	&	0.159	&	0.229 \\
Spread$_{t+1}$	&	0.169	&	0.242 \\
Depth$_{t+1}$	&	0.064	&	0.102 \\
Holdings$_{t+1}$	&	0.005	&	0.001 \\
Cash$_{t+1}$	&	0.005	&	0.001 \\
Trade Volume$_{t+1}$	&	0.113	&	0.178 \\
Traded Price$_{t+1}$	&	0.069	&	0.098 \\
Momentum($t+1$,$30$)	&	0.141	&	0.177 \\
Volatility $\sigma(t+1,30)$	&	0.001	&	0.002 \\
Reward $R(s_t, s_{t+1})$	&	0.0082	&	0.0004 \\ \hline
    \end{tabular}
    \label{tab:model_compare_stats}
    \vspace{-1ex}
\end{table}

% Any appendices should be placed after the list of references, beginning with the command \verb"\appendix" followed by the command \verb"\section" for each appendix title, e.g.
% \begin{verbatim}
% \appendix
% \section{This is the title of the first appendix}
% \section{This is the title of the second appendix}
% \end{verbatim}
% produces:\medskip

% \noindent\textbf{Appendix A. This is the title of the first appendix}\medskip

% \noindent\textbf{Appendix B. This is the title of the second appendix}\medskip

% \noindent Subsections, equations, figures, tables, etc.\ within appendices will then be automatically numbered as appropriate. Some theorem-like environments may need to have their counters reset manually (e.g.\ if they are not numbered within sections in the main text). You can achieve this by using \verb"\numberwithin{remark}{section}" (for example) just after the \verb"\appendix" command.

% Please note that if the \verb"endfloat" package is used on a document containing appendices, the \verb"\processdelayedfloats" command must be included immediately before the \verb"\appendix" command in order to ensure that the floats in the main body of the text are numbered as such.

%\processdelayedfloats %%% See above for an explanation of why this command might be needed.

\appendix

\end{document}